%% file: main.tex
\documentclass[10pt,logo,copyright]{nvidiatechreport}
\input{packages}

\usepackage{pifont}

\usepackage[nameinlink]{cleveref}
\crefname{equation}{Eq.}{Eqs.}
\crefname{figure}{Fig.}{Figs.}
\crefname{section}{Sec.}{Sec.}
\crefname{appendix}{App.}{App.}
\crefname{table}{Tab.}{Tabs.}
\crefname{algorithm}{Algo}{Algo}
\crefname{thm}{Thm}{Thm}
\Crefname{thm}{Thm}{Thm}
\crefname{prop}{Prop}{Prop}

\definecolor{darkred}{rgb}{0.7, 0.0, 0.0}

\newcommand{\crefnames}[3]{%
  \@for\next:=#1\do{%
    \expandafter\crefname\expandafter{\next}{#2}{#3}%
  }%
}

\input{sections/customized}

\title{\oursname{}: Train your LLMs to Decompose Query and Search Sub-queries in Parallel with Reinforcement Learning}

\author{
    {Shu Zhao}$^{1,2}$\footnote[1]{Work was conducted during Shu's internship at NVIDIA.}\quad {Tan Yu}$^{1,\dagger}$\quad {Anbang Xu}$^1$ \quad {Japinder Singh}$^{1}$\quad {Aaditya Shukla}$^{1}$\quad {Rama Akkiraju}$^{1}$\\
    \normalsize \textsuperscript{1} NVIDIA \quad
    \normalsize \textsuperscript{2} Pennsylvania State University\\
    $^{\dagger}$ Project Lead \texttt{tayu@nvidia.com}
}

\begin{document}

\maketitle

\input{sections/0_abstract}
\abscontent
\input{sections/1_introduction}
\input{sections/2_related_work}
\input{sections/3_method}
\input{sections/4_experiments}
\input{sections/5_conclusion}

\section{Acknowledgments}
We acknowledge the NVIDIA Brev Team for providing access to GPU clusters and training resources. Their technical support and computational infrastructure were instrumental in enabling the large-scale experiments reported in this paper.

\bibliographystyle{plainnat}
\bibliography{main}

\clearpage
\input{sections/X_appendix}

\end{document}

%% file: packages.tex
\usepackage[authoryear,round]{natbib}

\usepackage[utf8]{inputenc} 
\usepackage[T1]{fontenc}    

\usepackage{parskip}        
\usepackage{url}            
\usepackage{booktabs}       
\usepackage{amsfonts}       
\usepackage{nicefrac}       
\usepackage{microtype}      
\usepackage{xcolor}         
\usepackage[dvipsnames]{xcolor} 
\usepackage{graphicx}
\usepackage{animate}        
\usepackage{subcaption}
\usepackage{tabularx}
\usepackage{makecell}
\usepackage{adjustbox}
\usepackage{setspace}
\newcolumntype{M}[1]{>{\centering\arraybackslash}m{#1}}
\usepackage{float}
\usepackage{tikz}
\usetikzlibrary{positioning,shapes,arrows}
\usepackage{amsmath,amsfonts,bm, bbm,leftindex}
\usepackage{multirow}
\usepackage{comment}
\usepackage{gensymb}
\usepackage{lipsum}
\usetikzlibrary{arrows.meta, positioning, fit}
\usepackage[para]{threeparttable}
\usepackage{tikz}
\usetikzlibrary{tikzmark}

\usepackage[most]{tcolorbox}
\usepackage{fancyvrb}
\usepackage{fvextra}
\usepackage{dashrule}



%% file: sections/customized.tex
\usepackage{amsmath}
\usepackage{amssymb}
\usepackage{enumitem}
\usepackage{graphicx}
\usepackage{lineno}
\usepackage{xspace}
\usepackage{algorithm}
\usepackage{algpseudocode}
\usepackage{array}
\usepackage{booktabs}
\usepackage{longtable}
\usepackage{arydshln}
\usepackage{multirow}
\usepackage{multicol}
\usepackage{caption}
\usepackage{cleveref}
\usepackage{xcolor}
\usepackage{colortbl}
\usepackage{tcolorbox}
\usepackage[symbol]{footmisc}
\usepackage{longtable}
\usepackage{array}
\usepackage{placeins}

\newcommand{\think}[1]{\textcolor{blue}{\texttt{<think>}} #1 \textcolor{blue}{\texttt{</think>}}}
\newcommand{\search}[1]{\textcolor{cyan}{\texttt{<search>}} #1 \textcolor{cyan}{\texttt{</search>}}}
\newcommand{\info}[1]{\textcolor{brown}{\texttt{<information>}} #1 \textcolor{brown}{\texttt{</information>}}}
\newcommand{\answer}[1]{\textcolor{purple}{\texttt{<answer>}} #1 \textcolor{purple}{\texttt{</answer>}}}
\newcommand{\se}{\mathcal{R}}

\definecolor{box_green}{RGB}{0, 180, 139}
\definecolor{box_red}{RGB}{239, 99, 75}

\definecolor{verylightgray}{rgb}{0.9,0.9,0.9}
\definecolor{lightblue}{rgb}{0.733,0.875,1.0}
\definecolor{nvidiagreen}{rgb}{0.463, 0.725, 0}
\definecolor{lightgreen}{rgb}{0.73, 0.86, 0.5}
\definecolor{forestgreen}{rgb}{0.208,0.667,0.235}
\definecolor{verylightblue}{rgb}{0.839,0.925,1.0}
\definecolor{veryverylightblue}{rgb}{0.92,0.96,1.0}

\newcommand{\ours}{\textit{\textbf{\textcolor{nvidiagreen}{ParallelSearch}}}}
\newcommand{\oursname}{ParallelSearch}

%% file: sections/0_abstract.tex
\begin{abstract}
Reasoning-augmented search agents such as Search-R1, trained via reinforcement learning with verifiable rewards (RLVR), demonstrate remarkable capabilities in multi-step information retrieval from external knowledge sources. These agents address the limitations of their parametric memory by dynamically gathering relevant facts to address complex reasoning tasks. However, existing approaches suffer from a fundamental architectural limitation: they process search queries strictly sequentially, even when handling inherently parallelizable and logically independent comparisons. This sequential bottleneck significantly constrains computational efficiency, particularly for queries that require multiple entity comparisons. To address this critical limitation, we propose \ours{}, a novel reinforcement learning framework that empowers large language models (LLMs) to recognize parallelizable query structures and execute multiple search operations concurrently. Our approach introduces dedicated reward functions that incentivize the identification of independent query components while preserving answer accuracy through jointly considering correctness, query decomposition quality, and parallel execution benefits. Comprehensive experiments demonstrate that \oursname{} outperforms state-of-the-art baselines by an average performance gain of $2.9\%$ across seven question-answering benchmarks. Notably, on parallelizable questions, our method achieves a $12.7\%$ performance improvement while requiring only $69.6\%$ of the LLM calls compared to sequential approaches. The project page is available at: \url{https://shuzhao.me/ParallelSearchProject/}.
\end{abstract}

%% file: sections/1_introduction.tex
\section{Introduction}
Large Language Models (LLMs) have demonstrated exceptional proficiency in complex reasoning tasks~\citep{gpt4,gemini,llama3,qwen25_arxiv24}, yet remain fundamentally constrained by their static parametric knowledge, which cannot access real-time information or domain-specific facts beyond their training cutoff~\citep{rag_survey}. To address this limitation, recent advances, like Search-R1~\citep{searchr1_arxiv25}, in reasoning-augmented search agents have emerged, augmenting LLMs with multi-step information retrieval capabilities that enable dynamic knowledge acquisition from external sources through reinforcement learning to strategically formulate search queries, decompose complex questions, and synthesize retrieved information into coherent responses.

However, despite these advances, current reasoning-augmented search agents suffer from a critical architectural inefficiency: they process all queries through a strictly sequential pipeline, treating each search operation as inherently dependent on its predecessors~\citep{searchr1_arxiv25,zerosearch_arxiv25,otc_arxiv25}. This sequential bottleneck persists even when handling parallelizable queries that contain logically independent components, as shown in \Cref{fig:teaser}. Consider the question: ``\texttt{Who is older, Claude Monet or Camille Pissarro?}'' Current search agents must first generate and execute a query for Claude Monet's birth date, analyze the results, then separately query for Camille Pissarro's birth date, before finally synthesizing both results. This workflow requires three LLM calls and two sequential search operations, resulting in computational overhead and latency. In production environments serving millions of queries, such inefficiencies cascade into prohibitive computational costs and degraded user experiences~\citep{retrieval_user_satisfaction_jasist10}, particularly as agents tackle complex tasks requiring multiple steps.

\input{figures/teaser}

Motivated by these limitations, \textbf{our key insight is that many complex queries can be naturally decomposed into independent sub-components,  which can be searched concurrently without compromising reasoning quality.} In the aforementioned example, retrieving birth dates for Claude Monet and Camille Pissarro involves no logical dependencies, and these operations can and should be executed in parallel. Therefore, we propose \ours{}, a novel reinforcement learning framework, to train LLMs to recognize such parallelizable patterns, decompose queries into independent subqueries within a single reasoning step, execute these subqueries concurrently, and synthesize results efficiently. This approach reduces the number of required LLM calls and enables parallel search execution, resulting in a significant reduction in both inference time and costs.

Our framework leverages Reinforcement Learning with Verifiable Rewards~\citep{r1_arxiv25,ppo_arxiv17} to teach LLMs when and how to identify parallelizable query structures, generate multiple sub-queries in a single pass, and aggregate results effectively. The primary technical challenge lies in training models to accurately detect query independence while avoiding over-parallelization that could introduce redundant searches or miss critical dependencies. We address this through two novel reward components that complement the standard answer correctness reward and format reward: (1) a decomposition reward that explicitly incentivizes correct identification of parallelizable patterns, and (2) a search efficiency reward that encourages single-round completion for parallelizable queries while penalizing unnecessary sequential processing.

The main contributions of our work are as follows:
\begin{itemize}
\item We identify and formalize the parallel query execution problem in reasoning-augmented search agents, demonstrating through empirical analysis that current sequential processing paradigms incur substantial computational overhead for parallelizable queries.
\item We propose \ours{}, a reinforcement learning framework employing verifiable rewards to train LLMs to recognize parallelizable queries, decompose complex questions into independent sub-queries, and execute concurrent searches while maintaining answer accuracy.
\item We demonstrate through extensive experiments across seven question-answering benchmarks that our method achieves a $2.9\%$ average performance improvement over SOTA baselines, with particularly notable gains of $12.7\%$ on parallelizable questions, while reducing LLM calls by $30.4\%$ and maintaining robust generalization to out-of-domain datasets.
\end{itemize}

%% file: figures/teaser.tex
\begin{figure*}[t]
    \centering
    \includegraphics[width=0.9\linewidth]{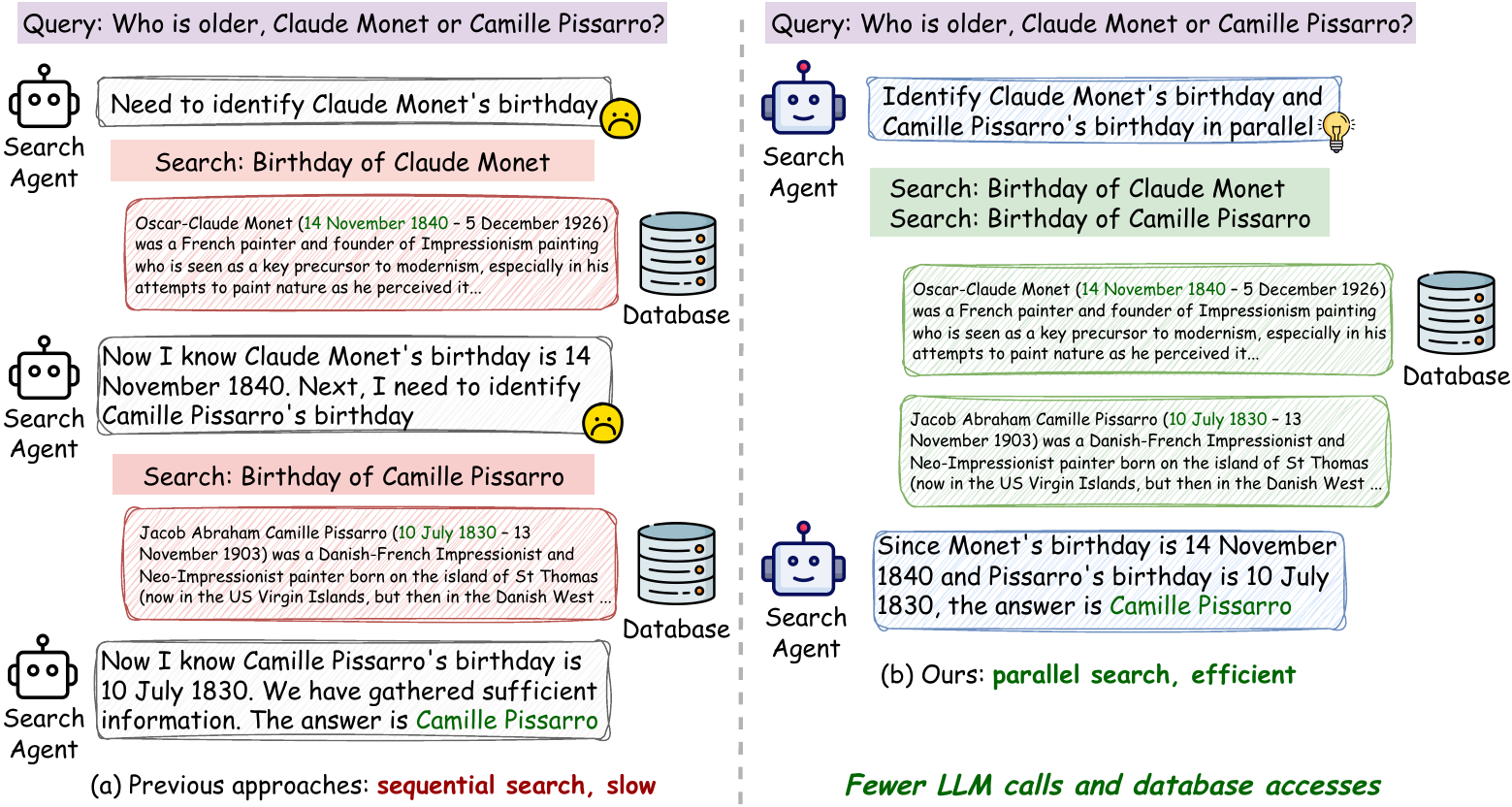}
    \caption{\textbf{Comparison of sequential versus parallel search strategies.} (a) Previous approaches process sub-queries sequentially, requiring multiple time-consuming reasoning steps and separate search rounds to gather information about each entity independently. (b) Our \ours{} method efficiently identifies independent sub-queries and executes them simultaneously in a single search round, significantly reducing both LLM inference calls and search latency while maintaining answer quality.}
    \label{fig:teaser}
\end{figure*}

%% file: sections/2_related_work.tex
\section{Related Work}

\subsection{Deep Search Agents}
While Large Language Models~(LLMs) exhibit impressive reasoning abilities~\citep{gpt4,gemini,llama3}, they suffer from insufficient domain-specific knowledge~\citep{llm_in_finance_survey} and hallucination issues~\citep{hallucination_survey}. To incorporate external data sources, a retriever can be integrated to provide external information to LLMs through two main approaches, including retrieval-augmented generation~(RAG)\citep{ralm_icml20} and utilizing retrievers as tools~\citep{toolformer_neurips23}. The RAG approaches~\citep{rag_survey} generally employ a retrieval phase followed by generation, wherein a retriever returns relevant content based on the input query. Nevertheless, this method can encounter difficulties with fetching irrelevant content~\citep{lc_meet_rag_iclr25}. The search-as-a-tool paradigm offers an alternative, enabling LLMs to engage with retrievers through prompting or fine-tuning. Methods like IRCoT~\citep{ircot_acl23} and ReAct~\citep{react_iclr23} employ prompting for iterative reasoning combined with retriever queries, whereas Toolformer~\citep{toolformer_neurips23} uses supervised fine-tuning to improve search functionality. Recent research, like Search-R1~\citep{searchr1_arxiv25}, employs reinforcement learning~\citep{r1_arxiv25} to train LLMs as agents for search tasks, achieving promising results. Follow-ups, like ZeroSearch~\citep{zerosearch_arxiv25}, O$^2$-Searcher~\citep{o2searcher_arxiv25}, MaskSearch~\citep{masksearch_arxiv25}, R-Search~\cite{rsearch_arxiv25}, and StepSearch~\cite{stepsearch_arxiv25}, further define fine-grained reward functions to improve the search performance. Despite these advances, existing search agents execute queries in a strictly sequential manner. Our \ours{} recognizes parallelizable query patterns, decomposes complex questions into independent sub-queries, and executes concurrent searches while maintaining accuracy.

\subsection{Reinforcement Learning}

Reinforcement learning~(RL) constitutes a decision-making paradigm where agents maximize long-term rewards by learning from environmental interactions and feedback signals~\citep{rl_intro_tnn98}. The application of RL to LLM fine-tuning began with \citet{instructgpt_neurips22}, who introduced RL from human feedback~(RLHF)~\citep{rlhf_survey_arxiv23}. In RLHF, human preference data~\citep{rewardbench_naaclf25} serves to construct a reward model that subsequently guides policy LLM optimization, predominantly through Proximal Policy Optimization~(PPO)~\citep{ppo_arxiv17}. The computational demands of PPO, which necessitates iterative LLM updates, pose substantial practical barriers. This has motivated the development of streamlined alternatives such as DPO~\citep{dpo_neurips23}, SimPO~\citep{simpo_neurips24}, and ORPO~\citep{orpo_emnlp24}, which bypass explicit reward modeling. While these methods reduce computational overhead, they struggle with off-policy limitations~\citep{iterative_rpo_neurips24} and often underperform relative to standard RL techniques. Novel solutions have emerged, including Group Relative Policy Optimization (GRPO)~\citep{r1_arxiv25}, which leverages group-based scoring to eliminate critic networks alongside RLOO~\citep{rloo_acl24}. The potential of RL for enhancing LLM capabilities in search engine utilization and reasoning tasks has received increasing attention. Our work extends verifiable rewards to parallel search planning, enabling RL training for learning complex parallelization patterns and addressing a critical gap where previous verifiable reward mechanisms cannot assess execution efficiency or dependency reasoning.

%% file: sections/3_method.tex
\input{tables/template}
\section{ParallelSearch}
\subsection{Preliminary}
\textbf{Search-R1}~\citep{searchr1_arxiv25} is an LLM-based search agent. It extends reasoning LLM~(\emph{e.g.}, deepseek-R1) by optimizing performance in interactions with a search engine. It is trained using reinforcement learning with verifiable rewards and performs multi-turn reasoning-search interactions. For each turn, the LLM leverages the reasoning capability to analyze the existing context, proposes a search query, and calls a search engine to discover the new information which would be attached to the context. The \textbf{iterative} process would terminate once there is sufficient information to generate the answer.
Specifically, Search-R1 follows a multi-turn reasoning-search loop: 
\begin{equation}
    \begin{aligned}
        \left(t_0, s_0, c_0, \cdots, t_{n-1}, s_{n-1}, c_{n-1}, t_n, a\right),
    \end{aligned}
\end{equation}
where $t$ denotes the think progress; $s$ is the search action; $c$ represents the retrieved contexts; $a$ is the final answer. The interactive multi-turn reasoning mechanism could address complex queries involving multiple entities/hops. Nevertheless, the step-by-step search reasoning process sometimes requires a large number of turns/rollouts to complete a task, resulting in high response latency, especially when the search engine is slow or LLM inference is time-consuming.

\noindent \textbf{Reinforcement learning} is widely used to optimize the LLM-based search agent. For instance, a commonly used objective is formulated as:
\begin{equation} \label{eq:rl-retriever}
\begin{aligned}
    \max_{\pi_\theta} \quad &\mathbb{E}_{x \sim \mathcal{D},\, y \sim \pi_{\theta}(\cdot \mid x; \se)}
    \left[ r_{\phi}(x, y) \right] \\
    - &\beta \mathbb{D}_{\text{KL}}\left[ \pi_{\theta}(y \mid x; \se) \,\|\, \pi_{\text{ref}}(y \mid x; \se) \right],
\end{aligned}
\end{equation}
where $R$ denotes the search engine, $x$ is the input query sampled from the data distribution $\mathcal{D}$,  $y$ represents the output sequence, $\pi_{\theta}$ denotes the policy LLM, $\pi_{\mathrm{ref}}$ is the reference LLM, and $r_{\phi}$ denotes the reward function. Search agents exploit two RL algorithms, including Proximal Policy Optimization (PPO)~\cite{ppo_arxiv17}  and Group Relative Policy Optimization (GRPO)~\cite{r1_arxiv25}, to compare their performance in retrieval-augmented reasoning.

\input{algorithms/alg_rlvr}
\subsection{ParallelSearch}

Observing the efficiency limitations of existing search agents, we propose \ours{} to decompose the query into sub-questions and encourage the use of parallel search tools when the original query is complex or involves multiple parts. The prompt template is illustrated in \Cref{tab:instruction}.

\subsubsection{Search-Reasoning Process}

In \Cref{alg:train}, we demonstrate the search-reasoning process of our proposed method. As the prompt template instructs the model to decompose the query into sub-questions delimited by $\#\#$, we use the delimiter to split the generated query surrounded by \texttt{<search>}\texttt{/<search>} into  multiple sub-questions. After that, we search the sub-questions in parallel, and aggregate the search results from all sub-questions as the context for reasoning and answer generation.  Benefited from executing the search engine in parallel, it decreases the number of turns/rollouts,  reducing the latency of the reasoning process. Moreover, compared to the sequential calling of the search engine in Search-R1, the parallel calling of the search engine also reduces the search overhead.

\subsection{Training Reward} 
The reward function in training consists of four components: outcome reward, decomposition reward, search count reward, and format reward, as illustrated in \Cref{fig:rlvr}.
\input{figures/rlvr}

\noindent \textbf{Outcome reward} evaluates the correctness of the final answer. To be specific, the correctness is measured by exact string matching (EM). Given the golden answer $a_{\textrm{gold}}$ and the predicted answer $a_{\mathrm{pred}}$, the outcome reward is defined by 
\begin{equation}
    r_{o} =  \textrm{EM}(a_{\textrm{gold}}, a_{\mathrm{pred}}),
\end{equation}
where $\textrm{EM}(x,y) = 1$ if $x$ and $y$ exactly match, and $0$ otherwise.

\noindent \textbf{Decomposition reward} encourages query decomposition for multi-hop and decomposable queries, while discouraging decomposition for non-decomposable multi-hop queries and single-hop queries. In the multi-hop training dataset, \emph{e.g.}, HotpotQA, the queries are classified into two categories: bridge and comparison. The comparison queries compare two entities, which are decomposable into two sub-queries, one for each entity. In contrast, the bridge queries require reasoning over bridging passages and are therefore non-decomposable. The decomposition reward is defined as
\begin{gather}
r_{d} = 
\begin{cases}
    \lambda_d           & \text{if } q \notin \mathcal{P} \land \neg \mathrm{D}(q), \\
    \alpha \lambda_d    & \text{if } q \in \mathcal{P} \land \mathrm{D}(q), \\
    0                   & \text{otherwise}, \\
\end{cases}
\end{gather}
where $q \in \mathcal{P}$ denotes the query $q$ is parallelizable; $\mathrm{D}(q)$ represents if the query is decomposed by the search agent; $\lambda_d$ is a positive constant controlling the weight of the decomposition reward, and $\alpha > 1$ is a positive constant to give a higher reward for decomposable queries considering that the decomposable queries are significantly fewer than the non-decomposable queries in the training dataset. 

\noindent \textbf{Search count reward} gives a penalty to the condition where there are a large number of search actions for decomposable queries, and implicitly encourages to decompose the query into sub-queries, and search the sub-queries in parallel within a single search action. On the other hand, for non-decomposable multihop queries, it penalizes the condition when there is only a search action. It is defined as 
\begin{gather}
r_{s} = 
\begin{cases}
    -\lambda_{s} \vert\textrm{count}_{ \langle \mathrm{search} \rangle} - 1\vert,           & \text{if } q \in \mathcal{P} \lor q \in \mathcal{S}\\
    -\lambda_{s} \vert\textrm{min}(\textrm{count}_{ \langle \mathrm{search} \rangle},2) - 2\vert,           & \text{otherwise}, \\
\end{cases}
\label{eq:search_count}
\end{gather}where $q \in \mathcal{S}$ represents the query is a single-hop question. $\textrm{count}_{\langle \mathrm{search} \rangle }$ denotes the total count of string $\langle \mathrm{search} \rangle$, \emph{i.e.}, the number of search turns, and $\lambda_s \in [0,1]$ is a predefined positive constant controlling the weight of the search count reward in the final reward. Note that, as shown in Eq.~(\ref{eq:search_count}), when $\mathrm{count}_{ \langle \mathrm{search} \rangle }=0$, the count reward is negative ($-\lambda_s$ or $-2\lambda_s$), giving a penalty to discourage the LLM from generating the answer based on its own knowledge without using search, in order to suppress the hallucination.

\noindent \textbf{Format reward} enourages LLM to follow the correct reasoning-action-observation format defined in \Cref{tab:instruction}, which is defined as 
\begin{gather}
r_{f} = 
\begin{cases}
    -\lambda_f            & \text{if } a_{\text{pred}} = a_{\text{gold}} \land \neg f_{\text{format}}(y), \\
    \lambda_f        & \text{if } a_{\text{pred}} \neq a_{\text{gold}} \land f_{\text{format}}(y), \\
    0              & \text{others} , \\
\end{cases}
\end{gather}
where $f_{\text{format}}(y) = 1$ if the response $y$ follows the format, otherwise $0$; $ \lambda_f$ is a positive constant. The final reward $r$ is defined as a summation of the above rewards:
\begin{equation}
    r = r_o + r_d + r_s + r_f.
\end{equation}

%% file: tables/template.tex
\begin{table}[t]
    \centering
    \caption{\textbf{Template for \ours{}.} \textcolor{red}{question} will be replaced with the specific question during training and inference.}\label{tab:instruction}
    \begin{tabular}{p{15.03cm}}
        \hline
        Answer the given question. \
        You must conduct reasoning inside \think{and} first every time you get new information. \
        After reasoning, if you find you lack some knowledge, you can call a search engine by \search{query}, and it will return the top searched results between \info{and}. \
        If the original query is complex or involves multiple parts, you are encouraged to decompose it into smaller sub-questions, separated by $\#\#$. For example: \search{sub-question 1 $\#\#$ sub-question 2}. \
        You can search as many times as you want. \
        If you find no further external knowledge needed, you can directly provide the answer inside \answer{and} without detailed illustrations. For example, \answer{xxx}.  Question: \textcolor{red}{question}.\\
        \hline
    \end{tabular}
\end{table}

%% file: algorithms/alg_rlvr.tex
\begin{algorithm}[t]
\caption{LLM Response Rollout with Multi-Turn  Search Engine Calls in Parallel}
\label{alg:llm_search}
\begin{algorithmic}[1]
\Require Input query \( x \), policy model \( \pi_{\theta} \), search engine \( \se \), maximum turns \( B \).
\Ensure Final response \( y \).

\State Initialize rollout sequence \( y \gets \emptyset \)
\State Initialize action count \( b \gets 0 \)

\While{\( b < B \)}
    \State Initialize current action LLM rollout sequence \( y_b \gets \emptyset \) 
    \While{True}
    \State Generate response token \( y_t \sim \pi_{\theta}(\cdot \mid x, y + y_b) \)
    \State Append \( y_t \) to rollout sequence \( y_b \gets y_b + y_t \)
    \If{\( y_t \) in [\textcolor{cyan}{\texttt{</search>}}, \textcolor{purple}{\texttt{</answer>}}, \texttt{<eos>}]}
        break
    \EndIf
    \EndWhile

    \State \( y  \gets  y + y_b \)
    \If{\search{} detected in \( y_b \)}
        \State Extract the concatenated sub-queries \( Q \gets \text{Parse}(y_b, \search{,} ) \)
        \State Split the concatenated sub-queries into individual sub-queries  \( \{q_i\}_{i=1}^N \gets \text{Split}(Q, \text{\#\#}) \)
        \State Retrieve search results in parallel \( \{d_i = \se(q_i)\}_{i=1}^n \)
        \State Insert $\{d\}_{i=1}^N$ into rollout \( y  \gets  y + \info{[d_1,\cdots,d_n]}  \)
    \ElsIf{\answer{} detected in \( y_b \)}
        \State \textbf{return} final generated response \( y \)
    \Else
        \State Ask for rethink \( y  \gets  y + \) ``My action is not correct. Let me rethink.''
    \EndIf

    \State Increment action count \( b \gets b + 1 \)
\EndWhile

\State \textbf{return} final generated response \( y \)
\end{algorithmic}
\label{alg:train}
\end{algorithm}

%% file: figures/rlvr.tex
\begin{figure}[t]
    \centering
    \includegraphics[width=0.8\linewidth]{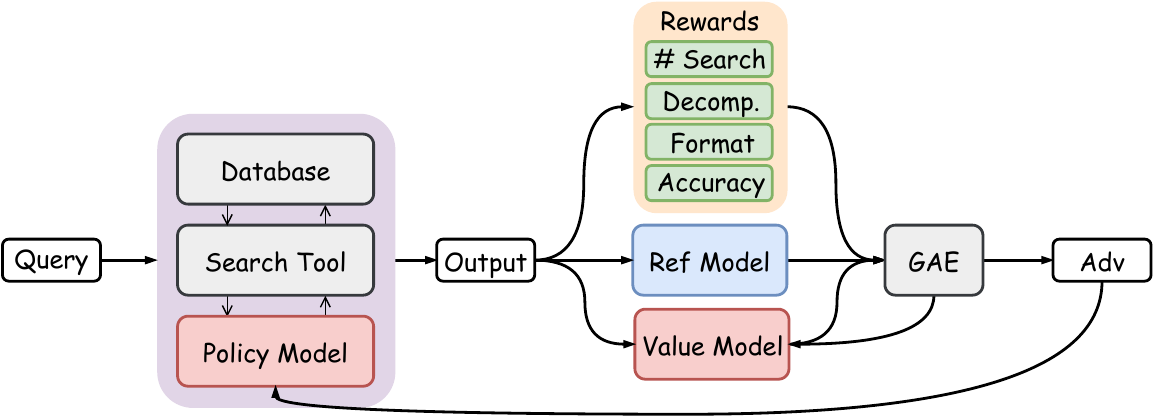}
    \caption{\textbf{Reinforcement learning with verifiable rewards for training \ours{}.} The reward function combines multiple components, including search efficiency, question decomposition, format adherence, and answer accuracy, to optimize correctness and parallel execution capabilities.}
    \label{fig:rlvr}
    \vspace{-3mm}
\end{figure}

%% file: sections/4_experiments.tex
\section{Results and Analysis}

\subsection{Experimental Setup}

\textbf{Datasets and Evaluation Metrics} \quad We select seven benchmark datasets that encompass a diverse range of search with reasoning challenges. These datasets are categorized as follows: (1) General Question Answering: NQ~\citep{nq_tacl19}, TriviaQA~\citep{triviaqa_acl17}, and PopQA~\citep{popqa_acl23}. (2) Multi-Hop Question Answering: HotpotQA~\citep{hotpotqa_emnlp18}, 2WikiMultiHopQA~\citep{2wikimultihopqa_coling20}, Musique~\citep{musique_tacl22}, and Bamboogle~\citep{bamboogle_emnlpf23}. Following~\citet{searchr1_arxiv25}, we merge the training sets of NQ and HotpotQA as the training data and conduct evaluations on the validation or test sets, using Exact Match~(EM) as the evaluation metric to assess both in-domain and out-of-domain performance.

\textbf{Baseline Methods} \quad We employ various baselines to evaluate our proposed \ours{}, including R1 without search engine~\citep{r1_arxiv25}, Search-R1~\citep{searchr1_arxiv25}, ZeroSearch~\citep{zerosearch_arxiv25}, StepSearch~\citep{stepsearch_arxiv25}, and OTC~\citep{otc_arxiv25}.

\textbf{Implementation Details} \quad We conduct experiments using Qwen-2.5-7B~(Base/Instruct) models~\citep{qwen25_arxiv24} as the backbone of the search agent, E5~\citep{e5_arxiv22} as the embedding model, and 2018 Wikipedia dump~\citep{dpr_emnlp20} as the corpus. We set the number of retrieved passages to $3$ across all retrieval-based methods following~\citet{searchr1_arxiv25}. All experiments are conducted on $8$ NVIDIA H100 GPUs. Detailed parameters used in experiments can be found in \Cref{sec:implementatoin_details}.

\subsection{Main Results}
\input{tables/main}
\input{tables/parallel}
\Cref{tab:main} presents the main results comparing \ours{} to baselines across seven benchmarks and four model configurations. From the results, we observe that: \textbf{(1) Our method consistently outperforms all baselines.} We achieve $2.9\%$ absolute improvement compared to the SOTA baselines, demonstrating the effectiveness of our proposed method. \textbf{(2) The improvement of the \texttt{instruct} model is larger than the \texttt{base} model.} We achieve absolute improvements of $3.4\%$ and $2.4\%$ using the \texttt{Instruct} and \texttt{Base} models, respectively. Interestingly, the performance of the model trained on \texttt{Base} outperforms that of the model trained on \texttt{Instruct} by $3.3\%$. This suggests that the parallel search capabilities are more effectively utilized by models that have not been instruction-tuned, which may preserve more abilities from the pre-training. \textbf{(3) Our method is better for complex multi-hop questions.} On four multi-hop benchmarks, our method outperforms the SOTA baseline by $3.01\%$, showing that our method learns the strategy for handling sequential and parallel questions. On single-hop questions, our method achieves comparable performance to \texttt{ZeroSearch}, which trains an LLM as a search engine simulator to generate pseudo-context information. Compared to \texttt{Search-R1}, our method outperforms it by $3.95\%$. \texttt{OTC} proposes a reward function to give a penalty to trajectories with longer interactions. However, \texttt{OTC} may lead to reward hacking: the agent is prone to not calling the search engine even when the information is not sufficient, reducing the performance by $6.0\%$ compared to our method.

\subsection{Parallel Behavior Analysis}
To evaluate the effectiveness and efficiency of our proposed method, we create subsets, including 2wiki-par, HotpotQA-par, and MultihopRAG-par, which only contain parallelizable questions from HotpotQA~\citep{hotpotqa_emnlp18}, 2WikiMultiHopQA~\citep{2wikimultihopqa_coling20}, and MultihopRAG~\citep{multihoprag_colm24}, respectively. Details about the creation of subsets can be found in \Cref{sec:par_seq_split}.

\input{figures/num_turns_cdf}
\input{figures/response_length}
\input{figures/reward_design}
\input{tables/decomp_score}
\textbf{Parallel searching reduces the number of LLM calls.} \Cref{tab:2wiki} shows the number of turns~(\texttt{\# Turns}) that the search agent interacts with LLMs. Compared to Search-R1, which retrieves information via sequential query generation, \ours{} significantly reduces the number of LLM calls ($2.34$ vs. $3.36$). It demonstrates that our method decomposes parallelizable questions into sub-queries and gathers all the required information using fewer turns. To further analyze the behavior of question decomposition, we report the cumulative distribution functions in \Cref{fig:num_turns_cdf}. The results show that our method efficiently responds to a question within $2$ turns, whereas Search-R1 requires $3-6$ turns. Note that the model autonomously determines parallelizability based on learned patterns during testing time, demonstrating its generalization ability.

\textbf{RL improves the inference efficiency.} 
We report the results of using the parallel prompt with Search-R1 as shown in \Cref{tab:2wiki} and \Cref{fig:num_turns_cdf}. The decomposition ratio of \texttt{Search-R1+Parallel Prompt} shows that providing the parallel prompt to Search-R1 can induce the query decomposition ability. However, the number of turns is not reduced, and we believe that \texttt{Search-R1} cannot correctly follow the parallel instruction. But in some cases, we observe that the inference time is lower than the baseline. To further analyze the inference latency, we report the response length in \Cref{fig:response_length}. \texttt{Search-R1} generates more tokens to support reasoning over retrieved documents. In contrast, \ours{} learns efficient reasoning given retrievals, significantly reducing the response length, suggesting that our method could reduce deployment costs and increase throughput, which are essential for real-world applications. Note that HotpotQA-par is an in-domain dataset, while 2WikiMultiHopQA-par and MultihopRAG-par are out-of-domain datasets. The results illustrate that the parallel decomposition ability can be applied across these datasets.

\textbf{Parallel searching outperforms sequential searching on parallelizable questions.} In \Cref{tab:2wiki}, \ours{} achieves an average improvement of $3.05\%$ compared to the baseline method, demonstrating its effectiveness on queries that inherently contain independent sub-components. Moreover, our method outperforms the baseline by $4.27\%$ on out-of-domain datasets, with consistent improvements across 2WikiMultiHopQA ($+4.5\%$) and MultihopRAG ($+3.2\%$), indicating that the learned parallelization patterns transfer effectively to new domains. 

\subsection{Reward Design Analysis}
\textbf{Reward function design is important.} Recent work~\citep{spurious_arxiv25} demonstrates that LLMs can still be improved, even when provided with noisy or even random rewards. \citet{contamination_arxiv25} suggests that this is likely due to a data contamination issue, which Qwen models may encounter during pre-training, leading to mathematical and coding problems. In our experiments, we find that the design of the reward function is critical and can significantly affect performance. We argue that this is due to the multi-turn interaction with LLMs, such as search, which is not pre-trained. As shown in \Cref{tab:ablation_score}, $\lambda_d=0$ increases the number of turns and decreases the decomposition ratio. Additionally, $\lambda_s=0$ significantly increases the number of turns, showing that these scores effectively decompose the parallelizable questions and prevent excessive search. When setting $\lambda_d$ to $0.15$ and $\lambda_s$ to $0.35$, \ours{} achieves the best performance.

\textbf{$r_d$ and $r_s$ reduce the number of valid searches and response length.} As shown in \Cref{fig:reward_design} (a) and (b), our reward function maintains a lower number of valid searches and response length, although both are slightly increased to better gather and extract useful information. Note that the length of retrieval documents is included in the response length here. From \Cref{fig:reward_design} (c), our analysis reveals that the instruction-tuned model exhibits accelerated convergence and demonstrates superior initial training rewards. Nevertheless, both model variants achieve remarkably similar final reward values upon completion of training. The results indicate that reinforcement learning serves as an effective method for closing this performance differential over the training, ultimately allowing base models to attain equivalent levels of performance.

%% file: tables/main.tex
\begin{table*}[t]
    \centering
    \caption{\textbf{Exact Match (EM) scores across seven general and multi-hop question answering benchmarks.} The \colorbox{box_green!12}{\textcolor{box_green}{best}} and \colorbox{box_red!12}{\textcolor{box_red}{second best}} scores under each metric are highlighted in colors. $^\dagger/^\star$ denote in-domain/out-of-domain datasets. $^\ddagger$ represents methods trained on different training sets. The average performance of \ours{} consistently outperforms baselines.}\label{tab:main}
    \resizebox{0.9\linewidth}{!}{
    \begin{tabular}{lcccccccc}
        \toprule
        \multirow{2}{*}{\textbf{Methods}} & \multicolumn{3}{c}{\textbf{General QA}} & \multicolumn{4}{c}{\textbf{Multi-Hop QA}} & \multirow{2}{*}{\textbf{Avg.}} \\
        \cmidrule(r){2-4} \cmidrule(l){5-8}
         & \textbf{NQ$^\dagger$} & \textbf{TriviaQA$^\star$} & \textbf{PopQA$^\star$} & \textbf{HotpotQA$^\dagger$} & \textbf{2wiki$^\star$} & \textbf{Musique$^\star$} & \textbf{Bamboogle$^\star$} & \\
        
        \midrule
        \rowcolor{gray!20} \multicolumn{9}{l}{\textit{\textbf{Qwen2.5-7b-Instruct}}} \\
        R1 & 0.270 & 0.537 & 0.199 & 0.237 & 0.292 & 0.072 & 0.293 & 0.271  \\
        Search-R1 & 0.383 & 0.593 & 0.399 & 0.376 & 0.317 & 0.151 & 0.371 & 0.370 \\
        OTC & 0.389 & - & - & 0.381 & - & - & - & - \\
        $\text{ZeroSearch}^{\ddagger}$ &  \cellcolor{box_red!12} 0.436 & \cellcolor{box_green!12} 0.652 & \cellcolor{box_green!12} 0.488 & 0.346 & 0.352 & 0.184 & 0.278 &  \cellcolor{box_red!12} 0.391 \\
        $\text{StepSearch}^{\ddagger}$ & - & - & - &  \cellcolor{box_red!12} 0.386 & \cellcolor{box_red!12} 0.366 & \cellcolor{box_green!12} 0.226 & \cellcolor{box_red!12} 0.400 & - \\
        \hdashline
        \textbf{\ours{}}~(\textit{\textbf{Ours}}) & \cellcolor{box_green!12} 0.462 & \cellcolor{box_red!12} 0.628 & \cellcolor{box_red!12} 0.429 & \cellcolor{box_green!12} 0.429 & \cellcolor{box_green!12} 0.424 & \cellcolor{box_red!12} 0.197 & \cellcolor{box_green!12} 0.411 & \cellcolor{box_green!12} 0.425 \\
        \midrule
        \rowcolor{gray!20} \multicolumn{9}{l}{\textit{\textbf{Qwen2.5-7b-Base}}} \\
        R1 & 0.297 & 0.539 & 0.202 & 0.242 & 0.273 & 0.083 & 0.296 &  0.276 \\
        Search-R1  & \cellcolor{box_red!12} 0.488 & 0.644 & \cellcolor{box_red!12} 0.469 & \cellcolor{box_red!12} 0.436 & \cellcolor{box_red!12} 0.412 & 0.187 & 0.403 &  \cellcolor{box_red!12} 0.434 \\
        OTC & 0.446 & 0.623 & 0.425 & 0.383 & 0.363 & 0.152 & 0.391 & 0.398 \\
        $\text{ZeroSearch}^{\ddagger}$ & 0.424 & \cellcolor{box_green!12} 0.664 & \cellcolor{box_green!12} 0.604 & 0.320 & 0.340 & 0.180 & 0.333 & 0.409 \\
        $\text{StepSearch}^{\ddagger}$ & - & - & - & 0.380 & 0.385 & \cellcolor{box_red!12} 0.216 & \cellcolor{box_red!12} 0.467 & -\\
        \hdashline
        \textbf{\ours{}}~(\textit{\textbf{Ours}}) & \cellcolor{box_green!12} 0.492 & \cellcolor{box_red!12} 0.658 & 0.455 & \cellcolor{box_green!12} 0.457 &	\cellcolor{box_green!12} 0.452 &	\cellcolor{box_green!12} 0.229	& \cellcolor{box_green!12} 0.468	& \cellcolor{box_green!12} 0.458 \\
       
        \bottomrule
    \end{tabular}
    }
\end{table*}

%% file: tables/parallel.tex
\begin{table}[t]
\caption{\textbf{Parallel behavior analysis on parallelizable question subsets.} The \colorbox{box_green!12}{\textcolor{box_green}{best}} and \colorbox{box_red!12}{\textcolor{box_red}{second best}} scores under each metric are highlighted in colors. Performance metrics include Exact Match (EM), Decomposition Ratio (DR), inference time per question (seconds), and average number of turns. $^\dagger/^\star$ denote in-domain/out-of-domain datasets respectively.} \label{tab:2wiki}
\centering
\resizebox{0.7\linewidth}{!}{
\begin{tabular}{l|p{35pt}<{\centering}p{35pt}<{\centering}p{63pt}<{\centering}p{42pt}<{\centering}}
\toprule
\textbf{HotpotQA-par}$^\dagger$ & \textbf{EM}$\uparrow$ & \textbf{DR~(\%)}$\uparrow$ & \textbf{Inference (s)}$\downarrow$ & \textbf{\# Turns}$\downarrow$ \\ \midrule
\rowcolor{gray!20} \multicolumn{5}{l}{\textit{\textbf{Qwen2.5-7b-Instruct}}} \\
Search-R1 & 0.580 & - & 2.28 & \cellcolor{box_red!12} 2.37 \\
Search-R1+Parallel Prompt & \cellcolor{box_red!12} 0.620 & \cellcolor{box_red!12} 83.20 & \cellcolor{box_red!12} 2.22 & 2.46 \\
\ours{}~(\textit{\textbf{Ours}}) & \cellcolor{box_green!12} 0.656 & \cellcolor{box_green!12} 94.30 & \cellcolor{box_green!12} 2.03 & \cellcolor{box_green!12} 2.08 \\
\midrule
\rowcolor{gray!20} \multicolumn{5}{l}{\textit{\textbf{Qwen2.5-7b-Base}}} \\
Search-R1 & \cellcolor{box_red!12} 0.641 & - & \cellcolor{box_red!12} 2.59 & \cellcolor{box_red!12} 4.00 \\
Search-R1+Parallel Prompt & 0.634 & \cellcolor{box_red!12} 9.06 & 2.74 & 4.98 \\
\ours{}~(\textit{\textbf{Ours}}) & \cellcolor{box_green!12} 0.673 & \cellcolor{box_green!12} 97.03 & \cellcolor{box_green!12} 2.16 & \cellcolor{box_green!12} 2.13 \\
\bottomrule
\toprule
\textbf{2wiki-par}$^\star$ & \textbf{EM}$\uparrow$ & \textbf{DR~(\%)}$\uparrow$ & \textbf{Inference (s)}$\downarrow$ & \textbf{\# Turns}$\downarrow$ \\ \midrule
\rowcolor{gray!20} \multicolumn{5}{l}{\textit{\textbf{Qwen2.5-7b-Instruct}}} \\
Search-R1 & 0.476 & - & 2.37 & 3.05\\
Search-R1+Parallel Prompt & \cellcolor{box_red!12} 0.573 & \cellcolor{box_red!12} 97.59 & \cellcolor{box_red!12} 2.22 & \cellcolor{box_red!12} 2.64 \\
\ours{}~(\textit{\textbf{Ours}}) & \cellcolor{box_green!12} 0.650 & \cellcolor{box_green!12} 99.75 & \cellcolor{box_green!12} 2.06 & \cellcolor{box_green!12} 2.03 \\
\midrule
\rowcolor{gray!20} \multicolumn{5}{l}{\textit{\textbf{Qwen2.5-7b-Base}}} \\
Search-R1 & \cellcolor{box_red!12} 0.624 & - & \cellcolor{box_red!12} 2.60 & \cellcolor{box_red!12} 4.00 \\
Search-R1+Parallel Prompt & 0.617 & \cellcolor{box_red!12} 19.14 & 2.83 & 4.97 \\
\ours{}~(\textit{\textbf{Ours}}) & \cellcolor{box_green!12} 0.691 & \cellcolor{box_green!12} 99.86 & \cellcolor{box_green!12} 2.25 & \cellcolor{box_green!12} 2.34 \\
\bottomrule
\toprule
\textbf{MultihopRAG-par}$^\star$ & \textbf{EM}$\uparrow$ & \textbf{DR~(\%)}$\uparrow$ & \textbf{Inference (s)}$\downarrow$ & \textbf{\# Turns}$\downarrow$ \\ \midrule
\rowcolor{gray!20} \multicolumn{5}{l}{\textit{\textbf{Qwen2.5-7b-Instruct}}} \\
Search-R1 & \cellcolor{box_red!12} 0.509 & - & \cellcolor{box_red!12} 2.37 & \cellcolor{box_red!12} 2.73 \\
Search-R1+Parallel Prompt & 0.488 & \cellcolor{box_red!12} 59.77 & \cellcolor{box_green!12} 2.34 & 3.37 \\
\ours{}~(\textit{\textbf{Ours}}) & \cellcolor{box_green!12} 0.641 & \cellcolor{box_green!12} 88.67 & \cellcolor{box_red!12} 2.37 & \cellcolor{box_green!12} 2.57 \\
\midrule
\rowcolor{gray!20} \multicolumn{5}{l}{\textit{\textbf{Qwen2.5-7b-Base}}} \\
Search-R1 & 0.569 & - & \cellcolor{box_red!12} 2.42 & \cellcolor{box_red!12} 4.00 \\
Search-R1+ParallelPrompt & \cellcolor{box_green!12} 0.598 & \cellcolor{box_red!12} 19.66 & 2.55 & 4.95 \\
\ours{}~(\textit{\textbf{Ours}}) & \cellcolor{box_red!12} 0.573 & \cellcolor{box_green!12} 60.94 & \cellcolor{box_green!12} 2.24 & \cellcolor{box_green!12} 2.89 \\
\bottomrule
\end{tabular}
}
\end{table}

%% file: figures/num_turns_cdf.tex
\begin{figure*}[t]
    \centering
    \includegraphics[width=1.0\linewidth]{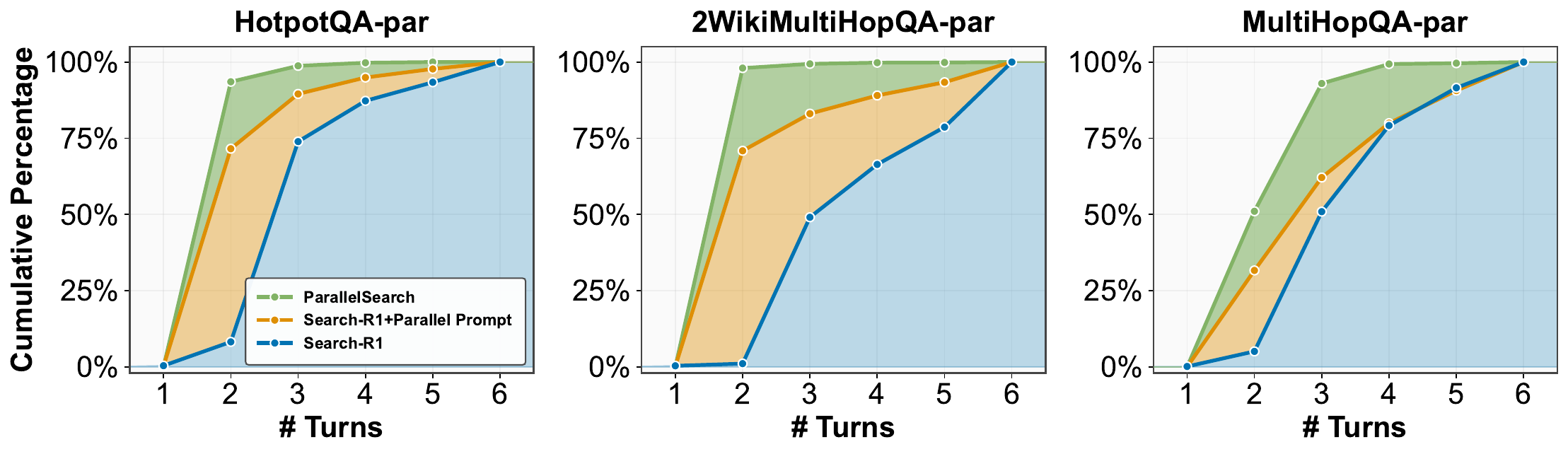}
    \caption{\textbf{Cumulative distribution of \# Turns across multi-hop QA datasets.} \ours{}~(green) consistently requires fewer turns compared to its counterparts, demonstrating improved efficiency through effective query parallelization.}
    \label{fig:num_turns_cdf}
\end{figure*}

%% file: figures/response_length.tex
\begin{figure}[t]
    \centering
    \includegraphics[width=0.6\linewidth]{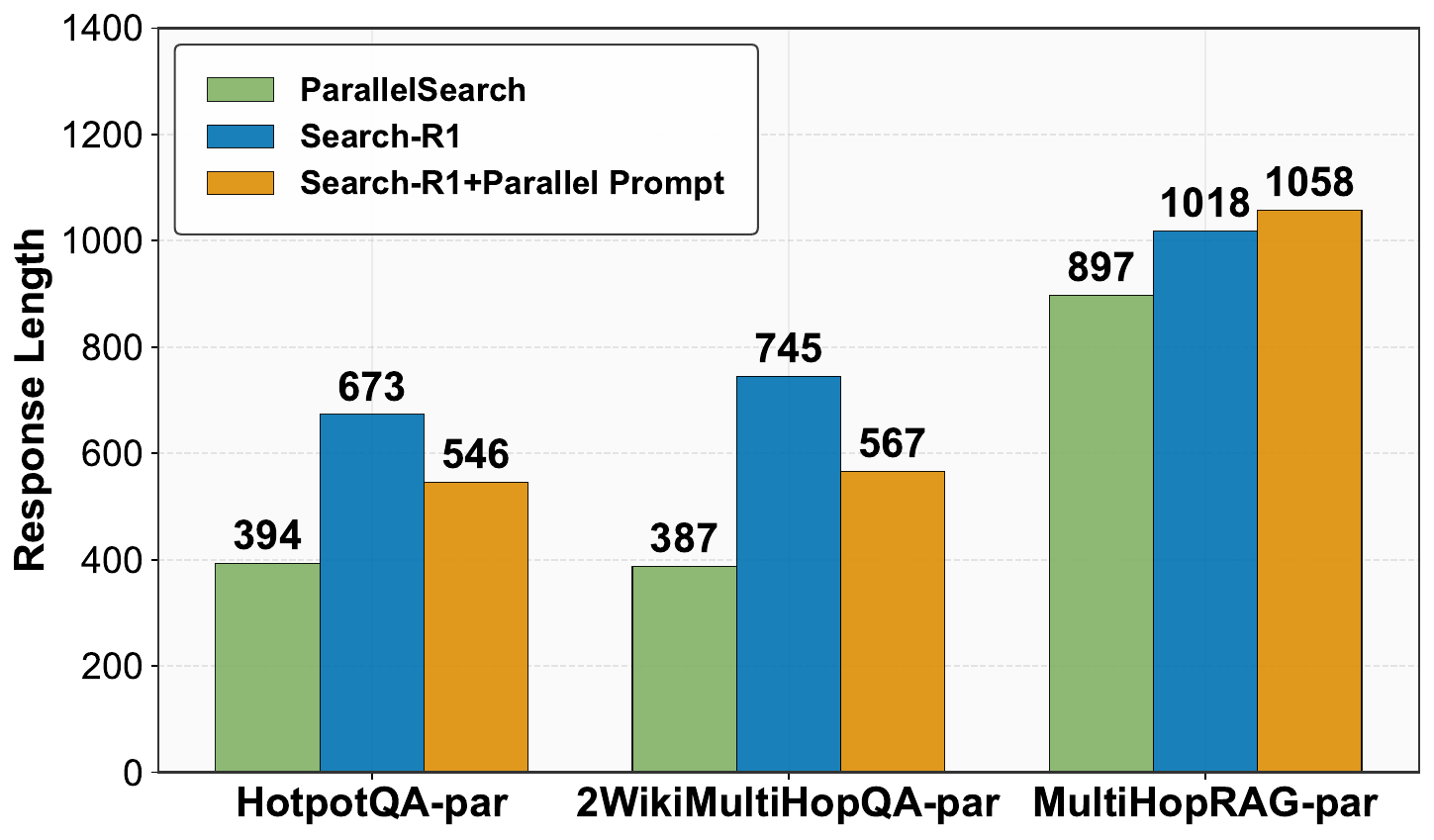}
    \captionof{figure}{\textbf{Average response length comparison across multi-hop QA datasets.} \ours{} achieves more concise responses while maintaining answer quality, showing efficient reasoning through parallel search strategies.}
    \label{fig:response_length}
    \vspace{-3mm}
\end{figure}

%% file: figures/reward_design.tex
\begin{figure*}[t]
    \centering
    \includegraphics[width=1.0\linewidth]{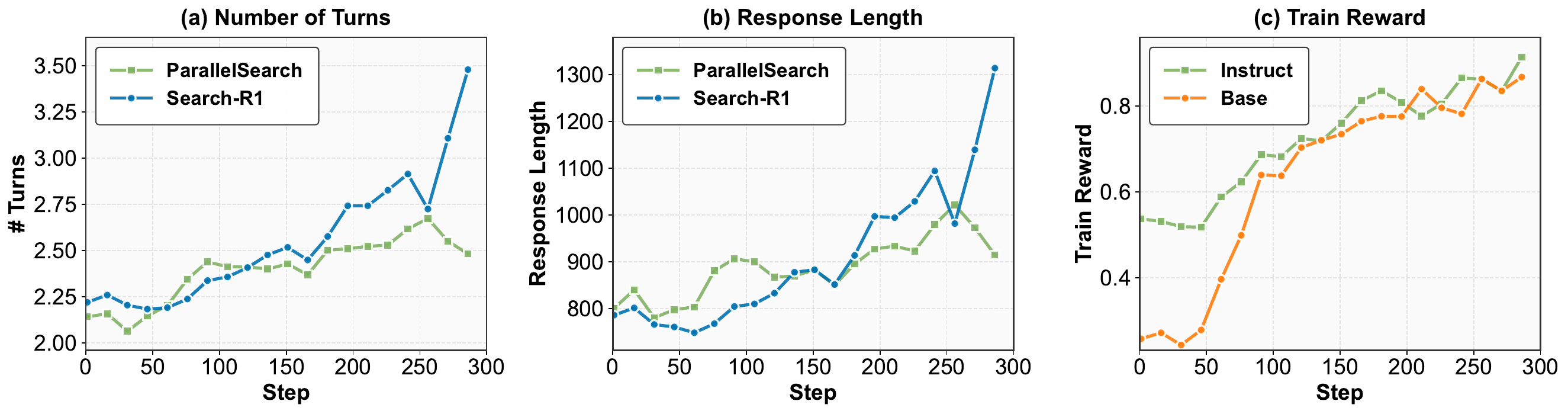}
    \caption{\textbf{Training dynamics during reinforcement learning}. (a) Number of turns remains at a low level. (b) Response length initially increases then stabilizes, balancing information gathering with conciseness. (c) Training reward curves show faster convergence for the instruction-tuned model compared to the base model, though both achieve similar final performance.}
    \label{fig:reward_design}
\end{figure*}

%% file: tables/decomp_score.tex
\begin{table}[t]
\centering
\caption{\textbf{Ablation study of decomposition reward $\lambda_d$ and search count reward $\lambda_s$ on parallelizable question subsets.}}\label{tab:ablation_score}
\begin{tabular}{cccccccc}
\toprule
\multirow{2}{*}{$\lambda_d$} & \multirow{2}{*}{$\lambda_s$} & \multicolumn{3}{c}{\textbf{HotpotQA-par}} & \multicolumn{3}{c}{\textbf{2wiki-par}} \\
\cmidrule(r){3-5} \cmidrule(l){6-8}
& & \textbf{EM} & \textbf{DR} & \textbf{\# Turns} & \textbf{EM} & \textbf{DR} & \textbf{\# Turns} \\
\midrule
0.00 & 0.35 & 0.636 & 90.94 & 2.22 & 0.610 & 98.33 & 2.13\\
\textbf{0.15} & \textbf{0.35} & 0.656 & 94.30 & 2.08 & 0.650 & 99.75 & 2.03\\
0.35 & 0.35 & 0.620 & 91.02 & 2.32 & 0.599 & 96.91 & 2.45\\
0.50 & 0.35 & 0.641 & 95.63 & 2.06 & 0.606 & 99.61 & 2.03\\
\hdashline
0.15 & 0.00 & 0.610 & 96.41 & 2.67 & 0.605 & 99.72 & 2.96\\
0.15 & 0.15 & 0.650 & 92.89 & 2.13 & 0.621 & 98.54 & 2.13 \\
\textbf{0.15} & \textbf{0.35} & 0.656 & 94.30 & 2.08 & 0.650 & 99.75 & 2.03\\
0.15 & 0.50 & 0.641 & 95.63 & 2.06 & 0.606 & 99.61 & 2.03\\
\bottomrule
\end{tabular}
\vspace{-3mm}
\end{table}

%% file: sections/5_conclusion.tex
\section{Conclusion}
We presented \ours{}, a novel reinforcement learning framework that addresses the critical efficiency bottleneck in reasoning-augmented search agents. By training LLMs to recognize and exploit query parallelizability, our approach enables concurrent execution of independent sub-queries while maintaining answer accuracy. Through carefully designed reward functions that jointly optimize for correctness, decomposition quality, and search efficiency, our method achieves a $2.9\%$ average performance improvement across seven benchmarks, with particularly notable gains of $12.7\%$ on parallelizable questions while reducing LLM calls by $30.4\%$. Our method demonstrates that architectural improvements in search agents need not come at the cost of additional model parameters or training data. Instead, teaching models when and how to parallelize through reinforcement learning offers a promising path toward more efficient and scalable information retrieval systems.

%% file: sections/x_appendix.tex
\appendix
\section{Experiment Setup}
\subsection{Datasets}
We select seven benchmark datasets that encompass a diverse range of search with reasoning challenges, including (1) General Question Answering: NQ~\citep{nq_tacl19}, TriviaQA~\citep{triviaqa_acl17}, and PopQA~\citep{popqa_acl23}. (2) Multi-Hop Question Answering: HotpotQA~\citep{hotpotqa_emnlp18}, 2WikiMultiHopQA~\citep{2wikimultihopqa_coling20}, Musique~\citep{musique_tacl22}, and Bamboogle~\citep{bamboogle_emnlpf23}. We extraly introduce MultiHopRAG~\citep{multihoprag_colm24} to evaluate the performance using parallelizable and sequential questions. For training, we merge the training sets of NQ and HotpotQA to construct a dataset for
\ours{} and other fine-tuning based baselines.

\textbf{Natural Questions~(NQ)} is a large-scale open-domain question answering dataset designed to advance natural language understanding capabilities. The dataset comprises $307$K training examples, $8$K development examples, and $8$K test examples, all derived from Wikipedia pages presented to human annotators. NQ supports two distinct tasks: long answer selection, where systems must identify relevant passages, and short answer selection, which requires the extraction of specific answer spans. Human performance benchmarks have been established at $87\%$ F1 for long-answer selection and $76\%$ F1 for short-answer selection, setting challenging targets for computational models. The dataset is available in two formats: a comprehensive $42$ GB version containing full HTML content, and a simplified $4$ GB version with extracted text for the training set. Released under the Creative Commons Share-Alike 3.0 license, Natural Questions represents a significant resource for researchers working on reading comprehension and question answering systems, with the explicit goal of developing models capable of matching human-level understanding.

\textbf{TriviaQA} is a large-scale reading comprehension dataset comprising over $65$K question-answer-evidence triples, designed to challenge existing question-answering systems. The dataset features $95$K question-answer pairs originally authored by trivia enthusiasts, paired with independently gathered evidence documents averaging six per question, providing high-quality distant supervision. TriviaQA distinguishes itself from other reading comprehension datasets through three key characteristics: (1) relatively complex, compositional questions that require sophisticated understanding, (2) considerable syntactic and lexical variability between questions and their corresponding answer-evidence sentences, and (3) the necessity for cross-sentence reasoning to identify correct answers. Initial baseline evaluations reveal a substantial performance gap, with a feature-based classifier achieving $23\%$ accuracy and state-of-the-art neural networks reaching $40\%$, both significantly below the $80\%$ benchmark for human performance. This considerable margin underscores TriviaQA's value as a challenging testbed for advancing research in reading comprehension and question answering.

\textbf{PopQA} is a large-scale, open-domain question-answering dataset specifically designed to evaluate language models' ability to memorize and recall factual knowledge across varying levels of entity popularity. The dataset comprises $14$K entity-centric QA pairs systematically generated by converting knowledge tuples from Wikidata into natural language questions using carefully crafted templates. Each question is derived from one of $16$ diverse relationship types and includes comprehensive annotations of the subject entity, object entity, and relationship type, along with Wikipedia monthly page view statistics to quantify entity popularity. Unlike existing QA datasets that predominantly feature high-popularity entities, PopQA intentionally encompasses entities across the popularity spectrum, with particular emphasis on long-tail entities. This design enables fine-grained analysis of how well language models encode world knowledge in their parameters, revealing that while scaling improves memorization of popular knowledge, models continue to struggle with less popular factual information, necessitating retrieval augmentation for effective performance on long-tail entities.

\textbf{HotpotQA} is a comprehensive question answering dataset containing $113$K Wikipedia-based question-answer pairs designed to advance complex reasoning capabilities in QA systems. The dataset addresses critical limitations in existing QA benchmarks through four distinctive features: (1) questions that necessitate finding and reasoning over multiple supporting documents, requiring genuine multi-hop reasoning; (2) diverse questions unconstrained by pre-existing knowledge bases or schemas, ensuring broad coverage of reasoning patterns; (3) sentence-level annotations of supporting facts that enable strong supervision during training and facilitate explainable predictions; and (4) a novel category of factoid comparison questions that evaluate systems' abilities to extract relevant facts and perform comparative analysis. Unlike traditional single-hop QA datasets, HotpotQA challenges models to perform sophisticated reasoning chains while maintaining interpretability through the provision of explicit supporting evidence. The provision of sentence-level supporting facts represents a significant advancement, enabling QA systems not only to arrive at correct answers but also to demonstrate their reasoning process, thereby addressing the growing need for explainable AI in question-answering tasks.

\textbf{2WikiMultiHopQA} is a multi-hop question answering dataset comprising $192,606$ examples that addresses critical limitations in existing multi-hop QA benchmarks by guaranteeing genuine multi-hop reasoning requirements. The dataset uniquely combines structured data from Wikidata and unstructured text from Wikipedia summaries, leveraging logical rules and carefully designed templates to generate four distinct question types: comparison (requiring quantitative or logical comparisons), inference (requiring understanding of logical rules), compositional (requiring sequential answering of sub-questions), and bridge comparison (combining bridge entity identification with comparative reasoning). A distinguishing feature of 2WikiMultiHopQA is the inclusion of evidence information, which provides complete reasoning paths from question to answer, serving the dual purposes of enabling comprehensive explanations for model predictions and facilitating rigorous evaluation of reasoning capabilities. The dataset exhibits diverse answer types, with the top categories being yes/no ($31.2\%$), dates ($16.9\%$), films ($13.5\%$), humans ($11.7\%$), and cities ($4.7\%$), encompassing a total of $708$ unique answer types. Through its methodical construction pipeline, 2WikiMultiHopQA ensures that multi-hop reasoning is genuinely required for each question, addressing previous concerns about pseudo multi-hop questions in existing datasets.

\textbf{MuSiQue} is a reading comprehension dataset designed to address the prevalent issue of shortcut-based reasoning in existing multi-hop benchmarks. The dataset employs a novel bottom-up construction methodology that systematically selects and composes pairs of single-hop questions from five existing datasets, ensuring that the resulting $2-4$ hop questions require genuine connected reasoning where each step critically depends on information from previous steps. MuSiQue is offered in two variants: MuSiQue-Answerable, containing $25$K answerable questions, and MuSiQue-Full, which doubles the dataset size by pairing each answerable question with a closely similar unanswerable counterpart. The task requires models to identify answers and supporting textbfs from contexts containing up to $20$ textbfs, with the additional challenge in MuSiQue-Full being first to determine answerability. Empirical evaluations demonstrate that MuSiQue presents substantial challenges compared to existing datasets, with a $3\times$ increase in the human-machine performance gap and a $30$-point drop in F1 score for single-hop models, confirming its effectiveness in preventing disconnected reasoning strategies and promoting genuine multi-hop comprehension.

\textbf{Bamboogle} is a manually curated, multi-hop question-answering dataset that addresses the limitations of template-based automatic generation found in larger datasets, such as MuSiQue, 2WikiMultiHop, and CC. Comprising $125$ carefully crafted $2$-hop questions, Bamboogle is constructed through a human-centered approach where annotators read random Wikipedia articles and formulate questions by connecting two unrelated facts about the article's topic. For instance, from the Voyager $2$ article, annotators might combine the facts that it was the first probe to approach Uranus and was launched on a Titan IIIE rocket to create the question ``What rocket was the first spacecraft that ever approached Uranus launched on?'' A unique quality control mechanism involves filtering questions through internet search engines, including only those that fail to produce correct featured snippets, thereby ensuring the questions represent genuinely challenging, multi-hop reasoning tasks that are not readily answerable from web searches. Despite its smaller scale, Bamboogle's diverse, non-templated questions provide a valuable testbed for evaluating systems' abilities to perform authentic complex question decomposition beyond pattern matching.

\textbf{MultiHopRAG} is a question answering dataset specifically designed to evaluate retrieval and reasoning capabilities in Retrieval-Augmented Generation~(RAG) pipelines, addressing the complex scenarios commonly encountered in real-world applications. The dataset comprises $2,556$ queries, distributed across four categories: inference queries ($31.92\%$), comparison queries ($33.49\%$), temporal queries ($22.81\%$), and null queries ($11.78\%$), with approximately $88\%$ of the queries being answerable from the knowledge base. Each query requires evidence from $2$ to $4$ documents, with $42\%$ requiring two pieces of evidence, $30\%$ requiring three, and $15\%$ requiring four, reflecting varying levels of reasoning complexity. The knowledge base comprises six distinct types of news articles, covering $609$ unique news items, with an average of $2,046$ tokens per article. A distinguishing feature of MultiHop-RAG is its incorporation of document metadata alongside textual content, enabling evaluation of systems' abilities to leverage both structured and unstructured information. The dataset exhibits considerable linguistic diversity, with queries beginning with various interrogative words including ``does'' ($27\%$), ``what'' ($15\%$), ``which'' ($15\%$), and ``who'' ($14\%$), ensuring comprehensive coverage of question types encountered in practical RAG applications.

\input{tables/par_seq_data_size}
\subsection{Parallelizable and Sequential Subsets} \label{sec:par_seq_split}
In multi-hop datasets, the questions can be grouped into parallelizable questions, sequential questions, and their combinations. To further evaluate the effectiveness and efficiency of our proposed \ours{}, we create subsets including parallelizable or sequential questions only. Specifically, we use the \texttt{comparison}, \texttt{comparison}, and \texttt{comparison\_query} in HotpotQA, 2WikiMultiHopQA, and MultihopRAG, respectively, to create the datasets with parallel questions only, denoting HotpotQA-par, 2wiki-par, and MultihopRAG-par. Then, we use \texttt{bridge} in HotpotQA, \texttt{inference} and \texttt{compositional} in 2WikiMultiHopQA, and \texttt{inference\_query} in MultihopRAG to construct 2wiki-seq, HotpotQA-seq, and MultihopRAG-seq, respectively. The data statistic is shown in \Cref{tab:par_seq_data_size}.

\subsection{Baselines}
We employ various baselines to evaluate our proposed \ours{}, including R1 without search engine~\citep{r1_arxiv25}, Search-R1~\citep{searchr1_arxiv25}, ZeroSearch~\citep{zerosearch_arxiv25}, StepSearch~\citep{stepsearch_arxiv25}, and OTC~\citep{otc_arxiv25}.

\textbf{R1 without search engine} directly prompts \texttt{DeepSeek-R1-Distill-Qwen-7B} to answer the given questions without using any search engine to fetch external data sources.

\textbf{Search-R1} introduces a reinforcement learning framework that trains large language models to interleave reasoning with search engine interactions autonomously. Unlike traditional retrieval-augmented generation approaches that perform single-round retrieval, Search-R1 enables LLMs to dynamically generate multiple search queries during step-by-step reasoning by learning to produce special tokens (\texttt{<search>} and \texttt{</search>}) that trigger search engine calls. The framework utilizes reinforcement learning algorithms (PPO or GRPO) to optimize the LLM's ability to determine when external knowledge is required, formulate relevant search queries, and integrate retrieved information into ongoing reasoning processes. To ensure stable training, Search-R1 implements retrieved token masking, which prevents the model from optimizing over externally retrieved content, and uses simple outcome-based rewards that assess only the correctness of final answers. This approach allows the model to learn effective search strategies through trial and error, developing capabilities for self-verification and iterative refinement without requiring explicit supervision on search behavior.

\textbf{ZeroSearch} presents a reinforcement learning framework that trains LLMs to develop search capabilities without requiring interaction with real search engines. The key innovation lies in using a lightweight fine-tuned LLM as a simulated search engine that can generate both useful and noisy documents in response to queries. During RL training, ZeroSearch employs a curriculum-based rollout strategy where the quality of generated documents is progressively degraded over time, exposing the policy model to increasingly challenging retrieval scenarios. This approach addresses two critical limitations of training with real search engines: the unpredictability of document quality and the prohibitive API costs associated with frequent search requests during RL rollouts. The framework is compatible with various RL algorithms (REINFORCE, PPO, GRPO). It demonstrates that even a 3B parameter simulation LLM can effectively incentivize search capabilities in the policy model, with larger simulators achieving performance comparable to or exceeding those of real search engines.

\textbf{StepSearch} introduces a reinforcement learning framework that trains LLMs as search agents through step-wise proximal policy optimization. Unlike previous search-RL methods that rely solely on global answer rewards, StepSearch augments PPO with fine-grained, token-level supervision that evaluates each search action based on information gain and redundancy penalties. The framework enforces a structured think-search-answer loop where the agent iteratively formulates search queries, retrieves documents from external sources, and incorporates the information into its reasoning process. At each search step, the method computes rewards by measuring the marginal information contribution of retrieved documents toward answering sub-questions (using cosine similarity over TF-IDF representations) while penalizing redundant retrievals that overlap with previously seen content. This step-wise supervision mechanism enables the model to decompose complex multi-hop queries into focused search subtasks and adapt its retrieval strategy dynamically. 

\subsection{Implementation Details} \label{sec:implementatoin_details}
We conduct experiments using Qwen-2.5-7B~(Base/Instruct) models~\citep{qwen25_arxiv24} as the backbone of the search agent, E5~\citep{e5_arxiv22} as the embedding model. For MultihopRAG, we use the data provided by the benchmark as the corpus; for other benchmarks, 2018 Wikipedia dump~\citep{dpr_emnlp20} is utilized as the corpus. To ensure fair comparison, we follow \citet{searchr1_arxiv25} and set the number of retrieved passages to $3$ across all retrieval-based methods. All experiments are conducted on $8$ NVIDIA H100 GPUs.

For PPO, we set the learning rate of the policy LLM and the value LLM  to $1$e$-6$ and $1$e$-5$, respectively. The number of training steps is $500$, with warm-up ratios of $0.285$ and $0.015$ for the policy and value LLMs, respectively. We set $\lambda = 1$ and $\gamma = 1$ for using Generalized Advantage Estimation~(GAE). The total batch size is $512$, with a mini-batch size of $256$ and a micro-batch size of $64$. The maximum sequence length is set to $4,096$ tokens. The maximum response length is $500$ and a maximum length is $500$ tokens for retrieved content. We enable gradient checkpointing and use Fully Sharded Data Parallel~(FSDP) with CPU offloading to optimize GPU memory usage. We utilize vLLM with a tensor parallel size of $1$, GPU memory utilization ratio of $0.6$, a temperature of $1.0$, a top-p value of $1.0$ for rollout. The KL divergence regularization coefficient $\beta$ and clip ratio $\epsilon$ are set to $0.001$ and $0.2$, respectively. 

\section{Sequential Search Analysis}
\input{tables/sequential}

ParallelSearch adaptively handles sequential questions while maintaining superior performance. Table~\ref{tab:sequential} presents results on sequential question subsets, which inherently require step-by-step information gathering where each search depends on previous results. Despite being optimized for parallel execution, \ours{} demonstrates remarkable adaptability.

\textbf{Consistent performance gains on sequential questions.} \ours{} achieves substantial improvements over Search-R1 across all sequential datasets, with average EM gains of 6.34\%, 1.30\%, and 8.86\% on HotpotQA-seq, 2wiki-seq, and MultihopRAG-seq respectively. This indicates that our reinforcement learning framework not only learns parallelization patterns but also develops more effective search strategies overall.

\textbf{Parallel prompting fails on sequential questions.} The Search-R1+Parallel Prompt baseline shows no improvement and often degrades performance on sequential questions, confirming that simply instructing models to parallelize is ineffective. In contrast, \ours{} learns through reinforcement learning to distinguish between parallelizable and sequential questions, applying the appropriate strategy dynamically.

\textbf{Robust generalization across models.} The performance gains are consistent across both Instruct and Base models, with the Base model showing particularly strong improvements (e.g., 40.61\% vs 37.86\% on HotpotQA-seq). This demonstrates that our method's benefits extend beyond instruction-tuned capabilities.

\section{Impact of Top-K Retrieval on Performance}
\input{figures/topk}
\Cref{fig:topk} investigates the impact of varying the number of retrieved passages (top-k) on the overall performance of \ours{}. The results demonstrate a consistent positive correlation between the number of retrieved documents and exact match accuracy across both 2WikiMultiHopQA and HotpotQA datasets. Specifically, increasing $k$ from $1$ to $10$ yields substantial improvements of approximately $6\%$ and $7\%$ for 2WikiMultiHopQA and HotpotQA respectively. The most significant gains occur when moving from top-$1$ to top-$3$ retrieval, with diminishing returns beyond $k=5$. This analysis validates our choice of $k=3$ in the main experiments, which strikes an optimal balance between computational efficiency and answer quality. The consistent improvement pattern across both datasets suggests that multi-hop reasoning benefits from access to multiple evidence sources, as parallel search operations can retrieve diverse documents that collectively contain the necessary information for complex question answering.

\section{Case Studies}
We conduct case studies to gain a deeper understanding of the behavior and capabilities of our \ours{} using different benchmarks and question types. The results are shown in the following tables.
\input{tables/case_hotpotqa_par_correct}
\input{tables/case_hotpotqa_par_wrong}
\input{tables/case_hotpotqa_seq_correct}
\input{tables/case_hotpotqa_seq_wrong}
\input{tables/case_2wiki_par_correct}
\input{tables/case_2wiki_par_wrong}
\input{tables/case_2wiki_seq_correct}
\input{tables/case_2wiki_seq_wrong}

%% file: tables/par_seq_data_size.tex
\begin{table}[t]
\centering
\caption{\centering\textbf{Data Statistic.}} \label{tab:par_seq_data_size}
\begin{tabular}{llr}
\toprule
\textbf{Dataset} & \textbf{Type} & \textbf{\# Samples} \\
\midrule
\multirow{2}{*}{HotpotQA} & Parallel & 1,487\\
& Sequential & 5,918 \\
\midrule
\multirow{2}{*}{2WikiMultiHopQA} & Parallel & 3,040 \\
& Sequential & 6,785 \\
\midrule    
\multirow{2}{*}{MultihopRAG} & Parallel & 856 \\
& Sequential & 816 \\
\bottomrule
\end{tabular}
\end{table}

%% file: tables/sequential.tex
\begin{table}[t]
\centering
\caption{\textbf{Behavior analysis on sequential question subsets.} The \colorbox{box_green!12}{\textcolor{box_green}{best}} and \colorbox{box_red!12}{\textcolor{box_red}{second best}} scores under each metric are highlighted in colors. Performance metrics include Exact Match (EM), inference time per question (seconds), and average number of turns. $^\dagger/^\star$ denote in-domain/out-of-domain datasets respectively.} \label{tab:sequential}
\resizebox{0.7\linewidth}{!}{
\begin{tabular}{l|p{35pt}<{\centering}p{63pt}<{\centering}p{42pt}<{\centering}}
\toprule
\textbf{HotpotQA-seq}$^\dagger$ & \textbf{EM}$\uparrow$ & \textbf{Inference (s)}$\downarrow$ & \textbf{\# Turns}$\downarrow$ \\ \midrule
\rowcolor{gray!20} \multicolumn{4}{l}{\textit{\textbf{Qwen2.5-7b-Instruct}}} \\
Search-R1 & \cellcolor{box_red!12}31.08 & \cellcolor{box_green!12}2.25 & \cellcolor{box_green!12}2.34 \\
Search-R1+Parallel Prompt & 31.05 & \cellcolor{box_red!12}2.28 & \cellcolor{box_red!12}3.33 \\
\ours{} & \cellcolor{box_green!12}37.42 & 2.50 & 3.55 \\
\midrule
\rowcolor{gray!20} \multicolumn{4}{l}{\textit{\textbf{Qwen2.5-7b-Base}}} \\
Search-R1-Base & \cellcolor{box_red!12}37.86 & \cellcolor{box_red!12}2.56 & \cellcolor{box_red!12}3.99 \\
Search-R1-Base+Parallel Prompt & 37.74 & 2.63 & 5.00 \\
\ours{} & \cellcolor{box_green!12}40.61 & \cellcolor{box_green!12}2.52 & \cellcolor{box_green!12}3.65 \\
\midrule
\midrule
\textbf{2wiki-seq}$^\star$ & \textbf{EM}$\uparrow$ & \textbf{Inference (s)}$\downarrow$ & \textbf{\# Turns}$\downarrow$ \\ \midrule
\rowcolor{gray!20} \multicolumn{4}{l}{\textit{\textbf{Qwen2.5-7b-Instruct}}} \\
Search-R1 & 21.15 & \cellcolor{box_green!12}2.30 & \cellcolor{box_green!12}2.70 \\
Search-R1+Parallel Prompt & \cellcolor{box_red!12}21.53 & \cellcolor{box_red!12}2.37 & 3.71 \\
\ours{} & \cellcolor{box_green!12}22.45 & 2.53 & \cellcolor{box_red!12}3.58 \\
\midrule
\rowcolor{gray!20} \multicolumn{4}{l}{\textit{\textbf{Qwen2.5-7b-Base}}} \\
Search-R1 & \cellcolor{box_red!12}25.84 & \cellcolor{box_green!12}2.56 & \cellcolor{box_green!12}3.99 \\
Search-R1+Parallel Prompt & 25.26 & 2.68 & 5.02 \\
\ours{} & \cellcolor{box_green!12}26.08 & \cellcolor{box_red!12}2.63 & \cellcolor{box_red!12}4.05 \\
\midrule
\midrule
\textbf{MultihopRAG-seq}$^\star$ & \textbf{EM}$\uparrow$ & \textbf{Inference (s)}$\downarrow$ & \textbf{\# Turns}$\downarrow$ \\ \midrule
\rowcolor{gray!20} \multicolumn{4}{l}{\textit{\textbf{Qwen2.5-7b-Instruct}}} \\
Search-R1 & \cellcolor{box_red!12}77.34 & \cellcolor{box_green!12}2.28 & \cellcolor{box_green!12}2.26\\
Search-R1+Parallel Prompt & 76.04 & \cellcolor{box_red!12}2.33 & \cellcolor{box_red!12}3.27 \\
\ours{} & \cellcolor{box_green!12}86.20 & 2.71 & 3.33 \\
\midrule
\rowcolor{gray!20} \multicolumn{4}{l}{\textit{\textbf{Qwen2.5-7b-Base}}} \\
Search-R1 & \cellcolor{box_red!12}87.76 & \cellcolor{box_green!12}2.40 & \cellcolor{box_red!12}3.97 \\
Search-R1+Parallel Prompt & 79.17 & \cellcolor{box_red!12}2.51 & 5.07 \\
\ours{} & \cellcolor{box_green!12}91.15 & 2.75 & \cellcolor{box_green!12}3.55 \\
\bottomrule
\end{tabular}
}
\end{table}

%% file: figures/topk.tex
\begin{figure}[t]
    \centering
    \includegraphics[width=0.7\linewidth]{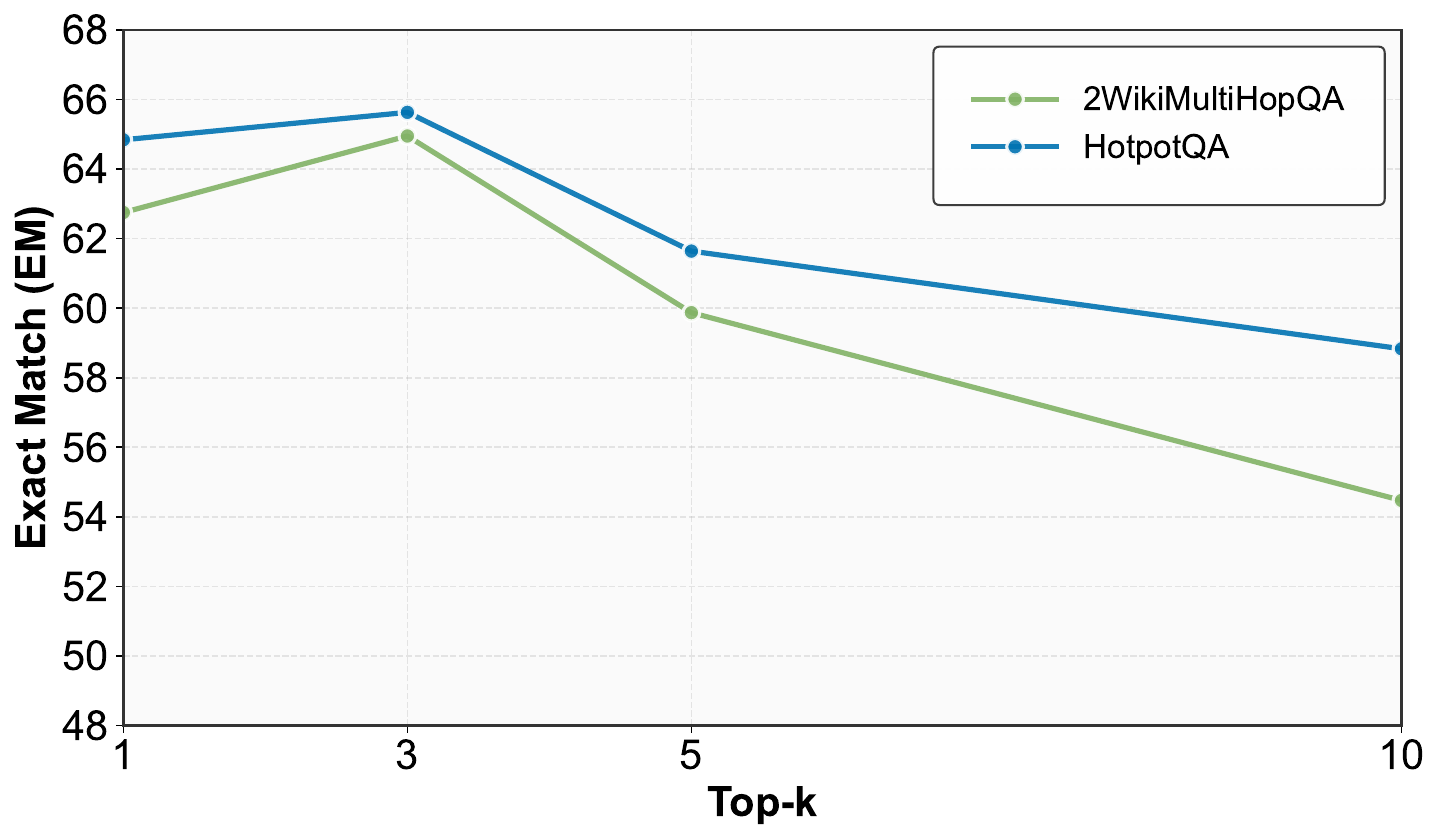}
    \caption{\textbf{Impact of top-k retrieval on exact match performance.} Performance comparison across different numbers of retrieved passages ($k \in \{1, 3, 5, 10\}$) on HotpotQA and 2WikiMultiHopQA datasets. Higher $k$ values consistently improve performance, with the most substantial gains occurring between $k=1$ and $k=3$, justifying our default setting of $k=3$ for all experiments.}
    \label{fig:topk}
\end{figure}

%% file: tables/case_hotpotqa_par_correct.tex
\begin{table*}[h]
    \centering
    \caption{\centering\textbf{Case study (correct) using the question from HotpotQA-par.}}\label{tab:case_hotpotqa_par_correct}
    \begin{tabular}{p{15.5cm}}
        \hline
        \textbf{Question}: Are the Laleli Mosque and Esma Sultan Mansion located in the same neighborhood? \\
        \hline
        \textbf{Ground Truth}: No \\
        \hline
        \textbf{\ours}: \\
        \think{I need to check the location of both the Laleli Mosque and Esma Sultan Mansion.} \\
\search{location of Laleli Mosque \#\# location of Esma Sultan Mansion} \\
\info{Doc 1(Title: Laleli Mosque) Laleli Mosque The Laleli Mosque () is an 18th-century Ottoman imperial mosque located in Laleli, Fatih, Istanbul, Turkey. The Laleli Mosque was built by Sultan Mustafa III from 17601763, designed in the baroque style by Ottoman imperial architect Mehmet Tahir Aa. The complex was destroyed by a fire in 1783 shortly after its completion and was immediately rebuilt. A fire in 1911 destroyed the madrasah, and subsequent road construction work destroyed many other auxiliary structures to the mosque. The mosque was built on a high terrace over a complex of vaulted shops, whose rents were intended to financially support the...\\
Doc 2(Title: Laleli Mosque) mosque complex. Underneath the mosque structure itself is a great hall, supported by eight enormous pillars with a fountain in the center. The mosque is oriented along a northwest-southeast axis, and has a rectangular courtyard about twice the size of the prayer hall to the northwest. This courtyard has a continuous arcade with eighteen domed bays and an ablution fountain in the center. The mosque itself has a brick and masonry base, with a masonry superstructure and octagonal drum supporting the dome. There are two minarets located at either end of the portico flanking the courtyard entrances. The Laleli Mosque...\\
Doc 3(Title: Laleli Mosque) of the narthex. Most of the structures of the Laleli Mosque kulliye have disappeared over the years, but octagonal domed trbe facing Ordu Street remains, and contains the graves of Mustafa III, his wife Mihrisah Sultan, son Selim III and daughters Hibetullah and Fatma Sultan and Mihrimah Sultan. The interior is decorated with znik tiles, and a band of calligraphy encircles the upper walls. The Tayyare Apartments were built in 1922 next to the mosque on the ground of 1911 burnt down Koska Madrasa, which belonged to the mosque's complex. The buildings were redeveloped into a five-star hotel. Laleli Mosque...\\
Doc 1(Title: Esma Sultan Mansion) Esma Sultan Mansion The Esma Sultan Mansion (), a historical yal () located at Bosphorus in Ortaky neighborhood of Istanbul, Turkey and named after its original owner Esma Sultan, is used today as a cultural center after being redeveloped. The three-storey brick building was designed by the renowned architect Sarkis Balyan and finished in 1875 next to Ortaky Mosque. It was presented to Esma Sultan, the daughter of Ottoman Sultan Abdulaziz, as a wedding gift. The mansion remained in the possession of the Ottoman dynasty until 1915. The building was subsequently used first as a tobacco warehouse and then as...\\
Doc 2(Title: Esma Sultan Mansion) guests in the ground floor and 600 guests in the first floor. The venue also hosts concerts of the Istanbul International Jazz Festival and the Istanbul International Music Festival. Esma Sultan Mansion The Esma Sultan Mansion (), a historical yal () located at Bosphorus in Ortaky neighborhood of Istanbul, Turkey and named after its original owner Esma Sultan, is used today as a cultural center after being redeveloped. The three-storey brick building was designed by the renowned architect Sarkis Balyan and finished in 1875 next to Ortaky Mosque. It was presented to Esma Sultan, the daughter of Ottoman Sultan Abdulaziz...\\
Doc 3(Title: Esma Sultan Mansion) a coal depot from 1920 until 1975, when it was destroyed by a fire. The ruin, consisting only of the outer walls of the building, was purchased in the early 1990s by The Marmara hotel chain. Following a renovation with additions designed by architects Haluk Sezgin and Philippe Robert, the mansion was opened in 2001 as a multipurpose event venue. Within the brick exterior, which was left as the original, a steel and glass structure is incorporated. According to the GAD Architecture group, a redesign by the architect Gkhan Avcolu was completed in 2005. The building includes a bar, a...
} \\
\hdashline
\think{Laleli Mosque is located in Laleli, Fatih, Istanbul, while Esma Sultan Mansion is located in Ortaky neighborhood of Istanbul. Therefore, they are not in the same neighborhood.} \\
\answer{No}\\
        \hline
    \end{tabular}
\end{table*}

%% file: tables/case_hotpotqa_par_wrong.tex
\begin{table*}[h]
    \centering
    \caption{\centering\textbf{Case study (wrong) using the question from HotpotQA-par.}}\label{tab:case_hotpotqa_par_wrong}
    \begin{tabular}{p{15.5cm}}
        \hline
        \textbf{Question}: Are both Cypress and Ajuga genera? \\
        \hline
        \textbf{Ground Truth}: No \\
        \hline
        \textbf{\ours}: \\
        \think{I need to check if both Cypress and Ajuga are genera.} \\
\search{Is Cypress a genus \#\# Is Ajuga a genus} \\
\info{Doc 1(Title: Cupressus) Cupressus Cupressus is one of several genera within the family Cupressaceae that have the common name cypress; for the others, see cypress. It is considered a polyphyletic group. Based on genetic and morphological analysis, the genus Cupressus is found in the subfamily Cupressoideae. The common name comes from Old French cipres and that from Latin cyparissus, which is the latinisation of the Greek. As currently treated, these cypresses are native to scattered localities in mainly warm temperate regions in the Northern Hemisphere, including western North America, Central America, northwest Africa, the Middle East, the Himalayas, southern China and...\\
Doc 2(Title: Taxodium) Taxodium Taxodium is a genus of one to three species (depending on taxonomic opinion) of extremely flood-tolerant conifers in the cypress family, Cupressaceae. The generic name is derived from the Latin word taxus, meaning yew, and the Greek word  (eidos), meaning similar to. Within the family, Taxodium is most closely related to Chinese swamp cypress (Glyptostrobus pensilis) and sugi (Cryptomeria japonica). Species of Taxodium occur in the southern part of the North American continent and are deciduous in the north and semi-evergreen to evergreen in the south. They are large trees, reaching tall and (exceptionally ) trunk diameter. The...\\
Doc 3(Title: Taxodium) during the Pliocene. Taxodium Taxodium is a genus of one to three species (depending on taxonomic opinion) of extremely flood-tolerant conifers in the cypress family, Cupressaceae. The generic name is derived from the Latin word taxus, meaning yew, and the Greek word  (eidos), meaning similar to. Within the family, Taxodium is most closely related to Chinese swamp cypress (Glyptostrobus pensilis) and sugi (Cryptomeria japonica). Species of Taxodium occur in the southern part of the North American continent and are deciduous in the north and semi-evergreen to evergreen in the south. They are large trees, reaching tall and (exceptionally )...\\
Doc 1(Title: Ajuga pyramidalis) Ajuga pyramidalis Ajuga pyramidalis, commonly known as pyramidal bugle, is a flowering plant of the genus Ajuga in the family Lamiaceae. It is a native plant in Europe. Pyramidal bugle is a perennial, herbaceous plant growing from about tall. At the base there is a rosette of stalked leaves which are significantly larger than the stem leaves. The stiff upright hairy stem is square and bears simple leaves growing in opposite pairs. They are ovate, hairy above and below and have a slightly wavy edge. The inflorescence has leaf-like bracts subtending the individual flowers. The bracts gradually get smaller towards...\\
Doc 2(Title: Ajuga decumbens) Ajuga decumbens Ajuga decumbens is a herbaceous flowering plant native to China, Japan and Korea. It is commonly found in lightly forested sunny areas, such as meadows and roadsides between 4002300 metres in altitude. This plant grows as a groundcover, and the leaf layer is usually no more than tall. It flowers between April and June. The purple to white erect flower stems can grow to tall, and are hermaphroditic. The corolla is straight, tubular and long. Petals are long. The leaves are purplish green and are wide to long. The entire plant is analgesic, decoagulant, depurative, febrifuge and haemostatic...\\
Doc 3(Title: Ajuga japonica) Ajuga japonica Ajuga japonica is a herbaceous flowering plant native to Japan. The species grows as a groundcover on the forest floor, usually near streams. It is often found in large clusters, due to its spreading and seeding habit. It grows along the floor with stolons, and is usually 8-20 centimetres high including the flowers. It flowers between late April and late May. The light purple flowers of Ajuga japonica are tubular and lip-shaped, being long. These flowers grow from the tips of the flower stems. The pinnate, heart-shaped leaves have serrated margins and are long. They are situated opposite...
} \\
\hdashline
\think{Both Cypress and Ajuga are identified as genera.} \\
\answer{Yes}\\
        \hline
    \end{tabular}
\end{table*}

%% file: tables/case_hotpotqa_seq_correct.tex
\begin{table*}[h]
    \centering
    \caption{\centering\textbf{Case study (correct) using the question from HotpotQA-seq.}}\label{tab:case_hotpotqa_seq_correct}
    \begin{tabular}{p{15.5cm}}
        \hline
        \textbf{Question}: 2014 S/S is the debut album of a South Korean boy group that was formed by who? \\
        \hline
        \textbf{Ground Truth}: YG Entertainment \\
        \hline
        \textbf{\ours}: \\
        \think{I need to know the boy group that debuted with the album 2014 S/S and who formed them. I will start by finding out the boy group.} \\
\search{who is the boy group that debuted with the album 2014 S/S} \\
\info{Doc 1(Title: 2014 S/S) Chart and Billboard's K-Pop Hot 100 chart. Internationally, 2014 S/S took the top spot in Billboard's World Album Chart. 2014 S/S 2014 S/S is the debut album of South Korean group WINNER. It was released on August 12, 2014 by the group's record label, YG Entertainment. The members were credited for writing the lyrics and composing the majority of the album's songs. The members produced the majority of the music for this album themselves, with the help of other producers such as Choice 37, Airplay, and others. The album was highlighted for incorporating elements generally absent from K-pop releases, including...\\
Doc 2(Title: 2014 S/S) 2014 S/S 2014 S/S is the debut album of South Korean group WINNER. It was released on August 12, 2014 by the group's record label, YG Entertainment. The members were credited for writing the lyrics and composing the majority of the album's songs. The members produced the majority of the music for this album themselves, with the help of other producers such as Choice 37, Airplay, and others. The album was highlighted for incorporating elements generally absent from K-pop releases, including hints of acoustic and alternative rock. The title track Empty became a hit in South Korea, topping the Gaon...\\
Doc 3(Title: Dark \& Wild) it stayed on the Billboard World Albums Charts for a total of 11 non-consecutive weeks. It also won Album Division at the 29th Golden Disk Awards in 2015. Dark \& Wild was the fourteenth bestselling album of South Korea Gaon Album Chart in 2014. All song credits are adapted from the Korea Music Copyright Association's database, unless otherwise noted. Dark \& Wild Dark \& Wild (stylized DARK \& WILD) is the debut studio album of South Korean boy group BTS. It was released by Big Hit Entertainment on August 19, 2014. The album contains fourteen tracks, with Danger as its...
} \\
\hdashline
\think{The boy group that debuted with the album 2014 S/S is WINNER. Now, I need to find out who formed them.} \\
\search{who formed the boy group WINNER} \\
\info{Doc 1(Title: Winner (band)) Winner (band) Winner (), stylized as WINNER, is a South Korean boy band formed in 2013 by YG Entertainment and debuted in 2014. It currently consists of four members, Jinwoo, Seunghoon, Mino and Seungyoon. A fifth original member, Taehyun, departed from the group in November 2016. The group was formed through the 2013 reality survival program WIN: Who is Next, where they competed as Team A against other trainees from YG Entertainment for the chance to debut as the first YG boy group in eight years, since the debut of Big Bang. At the conclusion of the program, the name...\\
Doc 2(Title: Winner (band)) in Taipei, Kuala Lumpur, Bangkok, Singapore, Manila, Jakarta, Hong Kong, United States, Canada, and more. Since the announcement, there have been releases of teasers to promote their world tour. Headlining Joint tours Opening act Winner (band) Winner (), stylized as WINNER, is a South Korean boy band formed in 2013 by YG Entertainment and debuted in 2014. It currently consists of four members, Jinwoo, Seunghoon, Mino and Seungyoon. A fifth original member, Taehyun, departed from the group in November 2016. The group was formed through the 2013 reality survival program WIN: Who is Next, where they competed as Team A...\\
Doc 3(Title: Kim Jin-woo (musician)) partake in the show as a fixed member. In January 2018, Jinwoo appeared on JTBC reality program Live a Good Life as a prisoner along with fellow label-mate Kwon Hyun-bin and Yoo Byung-jae. The reality show is a joint production of YG Entertainment and JTBC. List of awards and nominations received by Winner Individual Awards and Nominations: Kim Jin-woo (musician) Kim Jin-woo (born September 26, 1991), better known mononymously Jinwoo or Jinu, is a South Korean singer and actor. He is the lead vocalist of the South Korean boy group Winner, formed in 2013 by YG Entertainment through the survival...}\\
\hdashline
\think{WINNER is a South Korean boy group formed by YG Entertainment.}\\
\answer{YG Entertainment}\\
        \hline
    \end{tabular}
\end{table*}

%% file: tables/case_hotpotqa_seq_wrong.tex
\FloatBarrier
\begin{longtable}{p{15.5cm}}
    \caption{\centering\textbf{Case study (wrong) using the question from HotpotQA-seq.}}\label{tab:case_hotpotqa_seq_wrong}\\
    \hline
    \textbf{Question}: What screenwriter with credits for Evolution co-wrote a film starring Nicolas Cage and Ta Leoni? \\
    \hline
    \endfirsthead
    
    \multicolumn{1}{c}{\textit{(continued from previous page)}} \\
    \hline
    \endhead
    
    \hline
    \multicolumn{1}{c}{\textit{(continued on next page)}} \\
    \endfoot
    
    \hline
    \endlastfoot
    
    \textbf{Ground Truth}: David Weissman \\
    \hline
    \textbf{\ours}: \\
    \think{I need to identify the screenwriter who worked on Evolution and then check if they also co-wrote a film with Nicolas Cage and Ta Leoni. I will start by searching for the screenwriter of Evolution.} \\
    \search{screenwriter of Evolution} \\
    \info{Doc 1(Title: Evolution (2001 film)) Evolution (2001 film) Evolution is a 2001 American comic science fiction film directed by Ivan Reitman. It stars David Duchovny, Orlando Jones, Seann William Scott, Julianne Moore, and Ted Levine. It was released by DreamWorks in the United States and by Columbia Pictures internationally. The plot of the film follows college professor Ira Kane (David Duchovny) and geologist Harry Block (Orlando Jones), who investigate a meteor crash in Arizona. They discover that the meteor harbors extraterrestrial life, which is evolving very quickly into large, diverse and outlandish creatures. Evolution was based on a story by Don Jakoby, who turned it...\\
    Doc 2(Title: Evolution (TV series)) Evolution (TV series) Evolution is a 2001 documentary series by the American broadcaster Public Broadcasting Service (PBS) and WGBH on evolutionary biology, from the producers of NOVA. The spokespeople for the series were Jane Goodall (overall spokesperson), Kenneth R. Miller and Stephen Jay Gould (science spokespeople), Eugenie C. Scott (education spokesperson), Arthur Peacocke and Arnold Thomas (religious spokespeople). The series was narrated by the Irish actor Liam Neeson. The series was accompanied by a book by the popular science writer Carl Zimmer . An extensive website provides teaching resources for each episode's material, including The Mating Game, further looks at...\\
    Doc 3(Title: David Diamond (screenwriter)) its opening weekend, behind What Women Want and Cast Away, which opened at the top spot. Diamond and Weissman next wrote the sci-fi comedy Evolution, helmed by Ivan Reitman. Evolution was based on a story by Don Jakoby, who turned it into a screenplay along with Diamond and Weissman. The film was originally written as a serious horror science fiction film, until director Reitman re-wrote much of the script. A short-lived animated series, , loosely based on the film, was broadcast months after the film was released. They partnered with Wedding Crashers producer Andrew Panay on Old Dogs and When...
    } \\
    \hdashline
    \think{The screenwriter of Evolution is Don Jakoby, who turned the story into a screenplay with contributions from others. Now, I will search for a film starring Nicolas Cage and Ta Leoni and check if Don Jakoby co-wrote it.} \\
    \search{film starring Nicolas Cage and Ta Leoni} \\
    \info{Doc 1(Title: Nicolas Cage) film Nicolas Cage is a tormented man who hires a 15-year-old boy (played by Tye Sheridan) and protects him from his abusive father. The film premiered at the 70th Venice International Film Festival on August 30, 2013, with a subsequent screening at the 2013 Toronto International Film Festival. It was a box office flop, grossing only \$2.36 million from a \$4 million budget, but received critical acclaim from critics, who praised Cage's performance and Green's direction. On February 12, 2016, Cage was cast alongside Selma Blair and Anne Winters in Brian Taylor's horror comedy film, Mom and Dad. The film...\\
    Doc 2(Title: Tea Leoni) had the leading role in Deep Impact, a big-budget disaster film about a comet menacing Earth. The film received mixed reviews from critics, but it was a success at the box office, grossing \$349 million worldwide. She later had main roles in two other big budget movies: romantic comedy The Family Man (2000), co-starring alongside Nicolas Cage, and science fiction film Jurassic Park III (2001) as William H. Macy's character's ex-wife. In 2002, she starred as a film studio executive in Hollywood Ending, directed by Woody Allen and had a supporting role in the box office bomb crime drama People...\\
    Doc 3(Title: Nicolas Cage) coming to the attention of wider audiences with mainstream films, such as The Rock (1996), Face/Off (1997), Con Air (1997) and City of Angels (1998). He earned his second Academy Award nomination for his performance as Charlie and Donald Kaufman in Adaptation (2002). He also directed the film Sonny (2002), for which he was nominated for Grand Special Prize at Deauville Film Festival. Cage owns the production company Saturn Films and has produced films such as Shadow of the Vampire (2000) and The Life of David Gale (2003). He has also appeared in National Treasure (2004), Lord of War (2005)...}\\
    \hdashline
    \think{Since the search did not yield any results, I will conclude that either Don Jakoby did not co-write a film starring Nicolas Cage and Ta Leoni, or the film is not well-known or does not exist. Given the information available, I will deduce the answer based on known credits. Don Jakoby co-wrote Evolution, and a well-known film co-starring Nicolas Cage and Ta Leoni is The Family Man. However, Don Jakoby is not credited as a writer for The Family Man. I will assume the question might be referring to another film or a less known one. Since the direct answer is not found, I will provide the film name based on the known stars.}\\
    \answer{The Family Man}\\
\end{longtable}
\FloatBarrier

%% file: tables/case_2wiki_par_correct.tex
\begin{table*}[h]
    \centering
    \caption{\centering\textbf{Case study (correct) using the question from 2wiki-par.}}\label{tab:case_2wiki_par_correct}
    \begin{tabular}{p{15.5cm}}
        \hline
        \textbf{Question}: Which film came out first, Blind Shaft or The Mask Of Fu Manchu? \\
        \hline
        \textbf{Ground Truth}: The Mask Of Fu Manchu \\
        \hline
        \textbf{\ours}: \\
        \think{I need to search for the release dates of both films.} \\
\search{release date of Blind Shaft \#\# release date of The Mask Of Fu Manchu} \\
\info{Doc 1(Title: Blind Shaft) Blind Shaft Blind Shaft () is a 2003 film about a pair of brutal con artists operating in the illegal coal mines of present-day northern China. The film was written and directed by Li Yang (), and is based on Chinese writer Liu Qingbang's short novel Shen Mu (Sacred Wood). Most coal mines in China are worked by migrant workers who are forced to endure back-breaking, dangerous work in order to send money home. Some of them have additional schemes of their own. Song Jinming (played by Li Yixiang) and Tang Zhaoyang (Wang Shuangbao) are professional con artists, running an...\\
Doc 2(Title: My Blind Brother) it was reported that Starz had acquired distribution rights to the film. The film was scheduled to be released on September 23, 2016, in a limited release and through video on demand. It has a score of 59\% on Metacritic. The film has an 82\% Rotten Tomatoes score, based on 34 reviews. The site's consensus reads, My Blind Brother takes a refreshingly character-driven approach to its familiar sibling rivalry while adding a somewhat unexpected twist. Andrew Barker of Variety gave the film a positive review, writing that the film takes an ostensibly dark premise  a man of few qualities...\\
Doc 3(Title: Blindspotting) had to delay the film for an extended period of time. Principal photography finally began in June 2017 and lasted for 22 days, filming around Oakland. Blindspotting premiered at Sundance Film Festival on January 18, 2018, and was limited released in the United States on July 20, 2018, with a nation-wide release date of July 27, 2018. Blindspotting grossed \$332,500 in its opening weekend from 14 locations, including in Los Angeles, New York City, San Francisco, and Oakland, for a \$23,750 average per theater. It expanded to 513 theaters in its second weekend and made \$1.3 million. On review aggregation...\\
Doc 1(Title: The Mask of Fu Manchu) The Mask of Fu Manchu The Mask of Fu Manchu is a 1932 pre-Code adventure film directed by Charles Brabin. It was written by Irene Kuhn, Edgar Allan Woolf and John Willard based on the 1932 novel of the same name by Sax Rohmer. Starring Boris Karloff as Fu Manchu, and featuring Myrna Loy as his depraved daughter, the movie revolves around Fu Manchu's quest for the golden sword and mask of Genghis Khan. Lewis Stone plays his nemesis. Sir Denis Nayland Smith (Lewis Stone) of the British Secret Service warns Egyptologist Sir Lionel Barton (Lawrence Grant) that he must...\\
Doc 2(Title: Drums of Fu Manchu) It was one of the very rare films made under the Production Code to allow the villain to escape at the end. According to serial expert Alan Barbour, the Hays Office accepted the explanation that Fu Manchu always got away at the end of his novels to wreak further havoc in the next adventure, and Republic may well have been considering follow-ups. Drums of Fu Manchu's official release date is March 15, 1940, although this is actually the date the seventh chapter was made available to film exchanges. A 69-minute feature film version, created by editing the serial footage together...\\
Doc 3(Title: Fu Manchu) early talkie, The Mysterious Dr. Fu Manchu starring Warner Oland, best known for his later portrayal of Charlie Chan in the 1930s. Oland repeated the role in 1930's The Return of Dr. Fu Manchu and 1931's Daughter of the Dragon as well as in the short, Murder Will Out as part of the omnibus film, Paramount on Parade where the Devil Doctor confronts both Philo Vance and Sherlock Holmes. The most infamous incarnation of the character was MGM's The Mask of Fu Manchu (1932) starring Boris Karloff and Myrna Loy. The film's tone has long been considered racist and offensive...} \\
\hdashline
\think{Blind Shaft was released in 2003 and The Mask of Fu Manchu was released in 1932. Therefore, The Mask of Fu Manchu came out first.} \\
\answer{The Mask of Fu Manchu}\\
        \hline
    \end{tabular}
\end{table*}

%% file: tables/case_2wiki_par_wrong.tex
\begin{table*}[h]
    \centering
    \caption{\centering\textbf{Case study (wrong) using the question from 2wiki-par.}}\label{tab:case_2wiki_par_wrong}
    \begin{tabular}{p{15.5cm}}
        \hline
        \textbf{Question}: Are Fire In Hell and The Tiger: An Old Hunter'S Tale from the same country? \\
        \hline
        \textbf{Ground Truth}: yes \\
        \hline
        \textbf{\ours}: \\
        \think{I need to know the origin of both Fire In Hell and The Tiger: An Old Hunter's Tale.} \\
\search{origin of Fire In Hell \#\# origin of The Tiger: An Old Hunter's Tale} \\
\info{Doc 1(Title: The Origin of Fire) The Origin of Fire The Origin of Fire (), Op. 32, is a cantata composed by Finnish composer Jean Sibelius and premired on 9 April 1902 at the opening of the National Helsinki Theatre, conducted by the composer. It was later revised in 1910. The idea behind the cantata is taken from the Finnish national epic, the Kalevala. Some of the sketches for the piece can be related back to 1893/94. Originally scored for baritone, men's chorus and orchestra, the piece begins sombrely with the soloist narrating the first part of the story: The land of Kalevala is in darkness...\\
Doc 2(Title: Hell) Christian revelation, hell exists because evil people want it. They, not God, introduced evil to the human race. Members of The Church of Jesus Christ of Latter-day Saints (LDS Church) teach that hell is a state between death and resurrection, in which those spirits who did not repent while on earth must suffer for their own sins (Doctrine and Covenants 19:1517). In Islam, jahannam (in Arabic: ) (related to the Hebrew word gehinnom) is a place filled with blazing fire, boiling water, and a variety of other torments for those who have been condemned to it in the hereafter. After...\\
Doc 3(Title: Heresiarch) Heresiarch In Christian theology, a heresiarch (also hresiarch, according to the Oxford English Dictionary; from Greek: , hairesirkhs via the late Latin haeresiarcha) or arch-heretic is an originator of heretical doctrine, or the founder of a sect that sustains such a doctrine. In his Divine Comedy, Dante Alighieri represents the heresiarchs as being immured in tombs of fire in the Sixth Circle of Hell. In Cantos IX and X of the Inferno, Virgil describes the suffering these souls experience, saying Here are the Arch-Heretics, surrounded by every sect their followers... / Like with like is buried, and the monuments are...\\
Doc 1(Title: The Tiger: An Old Hunter's Tale) The Tiger: An Old Hunter's Tale The Tiger (also known as The Tiger: An Old Hunter's Tale, ; lit. Big Tiger) is a 2015 South Korean period drama film about a hunter prepared to kill the last tiger of Joseon. In Japanese-occupied Korea in 1925, Chun Man-duk, a revered hunter, lives with his teenage son, Seok, in a hut near Mount Jirisan. Following a tragic accident in which he killed his beloved wife, he has retired his rifle and become a humble herb gatherer limping amongst his cherished mountains. The Japanese governor-general overseeing the occupation gathers tiger pelts as a...\\
Doc 2(Title: Tiger) conflicts with humans. The tiger is among the most recognisable and popular of the world's charismatic megafauna. It featured prominently in ancient mythology and folklore and continues to be depicted in modern films and literature, appearing on many flags, coats of arms and as mascots for sporting teams. The tiger is the national animal of India, Bangladesh, Malaysia and South Korea. The Middle English tigre and Old English tigras (plural) derive from Old French tigre, from Latin tigris. This was a borrowing of Classical Greek  (transliterated as tigris, the modern species name), a foreign borrowing of unknown origin meaning...\\
Doc 3(Title: The Tiger: An Old Hunter's Tale) Man-duk reaches the top of the mountain and waits for the tiger. Soon following, the tiger appears. Man-duk and the tiger then seem the fight, with the former injuring the tiger. Man-duk provokes the tiger, asking why he stopped, and proceeds to take out a knife at the mountain's edge. The tiger pounces at him, and they both fall off the mountain together. The governor-general of the Japanese army asks the hunters what happened after the incident, and they relate to him a story about the Mountain Lord becoming a god. The governor-general comes to the conclusion that his army...} \\
\hdashline
\think{Fire In Hell does not have a clear country of origin mentioned, but The Tiger: An Old Hunter's Tale is from South Korea. Therefore, they are not from the same country.} \\
\answer{no}\\
        \hline
    \end{tabular}
\end{table*}

%% file: tables/case_2wiki_seq_correct.tex
\begin{table*}[h]
    \centering
    \caption{\centering\textbf{Case study (correct) using the question from 2wiki-seq.}}\label{tab:case_2wiki_seq_correct}
    \begin{tabular}{p{15.5cm}}
        \hline
        \textbf{Question}: When did John V, Prince Of Anhalt-Zerbst's father die? \\
        \hline
        \textbf{Ground Truth}: 12 June 1516 \\
        \hline
        \textbf{\ours}: \\
        \think{I need to know when John V, Prince of Anhalt-Zerbst's father died. I will start by finding out who his father was.} \\
\search{who is the father of John V, Prince of Anhalt-Zerbst} \\
\info{Doc 1(Title: John V, Prince of Anhalt-Zerbst) John V, Prince of Anhalt-Zerbst John V of Anhalt-Zerbst (Dessau, 4 September 1504  Zerbst, 4 February 1551), was a German prince of the House of Ascania and ruler of the principality of Anhalt-Dessau. From 1544, he assumed rule of the re-created principality of Anhalt-Zerbst. John was the second (but eldest surviving) son of Ernest I, Prince of Anhalt-Dessau, by his wife Margarete, daughter of Henry I, Duke of Mnsterberg-Oels, and granddaughter of George of Podbrady, King of Bohemia. Upon the death of his father in 1516, John and his brothers George III and Joachim I inherited Anhalt-Dessau as co-rulers...\\
Doc 2(Title: John VI, Prince of Anhalt-Zerbst) John VI, Prince of Anhalt-Zerbst John VI of Anhalt-Zerbst (Zerbst, 24 March 1621  Zerbst, 4 July 1667), was a German prince of the House of Ascania and ruler of the principality of Anhalt-Zerbst. He was the only son of Rudolph, Prince of Anhalt-Zerbst, by his second wife Magdalene, daughter of John VII, Count of Oldenburg. John succeeded his father in Anhalt-Zerbst at only four months of age; during his long minority, his paternal uncle Augustus of Anhalt-Pltzkau acted as regent in the principality. John's education was supervised primarily by his mother. Political instability caused by warfare during the Thirty...\\
Doc 3(Title: John V, Prince of Anhalt-Zerbst) union with the widowed daughter of the Elector of Brandenburg was a high honor for John, and he decided to celebrate the wedding with great pomp. But by that time, he was in poor health; finally, in 1544, he suffered a stroke. His relations with Margarete worsened during the following years; in 1550 John ordered the temporary arrest of his wife, but she fled. John and Margarete had six children: John V, Prince of Anhalt-Zerbst John V of Anhalt-Zerbst (Dessau, 4 September 1504  Zerbst, 4 February 1551), was a German prince of the House of Ascania and ruler of...
} \\
\hdashline
\think{John V, Prince of Anhalt-Zerbst's father is Ernest I, Prince of Anhalt-Dessau. Now, I will search when he died.} \\
\search{when did Ernest I, Prince of Anhalt-Dessau die} \\
\info{Doc 1(Title: Ernest I, Prince of Anhalt-Dessau) territories. The deaths of Sigismund III in 1487, George II in 1509, and Rudolph IV in 1510 without surviving male issue left Ernest as the sole ruler of Anhalt-Dessau until his death. In Cottbus on 20 January 1494 Ernest married Margarete (b. Breslau, 25 August 1473 - d. Dessau, 28 June 1530), daughter of Henry I, Duke of Mnsterberg-Oels, and granddaughter of George of Podbrady, King of Bohemia. They had four sons: Ernest I, Prince of Anhalt-Dessau Ernest I, Prince of Anhalt-Dessau (died Dessau, 12 June 1516), was a German prince of the House of Ascania and ruler of the...\\
Doc 2(Title: Ernest I, Prince of Anhalt-Dessau) Ernest I, Prince of Anhalt-Dessau Ernest I, Prince of Anhalt-Dessau (died Dessau, 12 June 1516), was a German prince of the House of Ascania and ruler of the principality of Anhalt-Dessau. He was the second son of George I, Prince of Anhalt-Dessau, yet the first born by his fourth wife Anna, daughter of Albert VIII, Count of Lindow-Ruppin. In 1473, after the death of his father, Ernest inherited the principality of Anhalt-Dessau alongside his younger brothers George II, Sigismund III, and Rudolph IV. Following the family law of the House of Ascania, the accession took place without any division of...\\
Doc 3(Title: Joachim Ernest, Prince of Anhalt) Prince Joachim Ernest died in Dessau on 6 December 1586. Since succession in the territories of Anhalt was not governed by the rules of primogeniture, Joachim Ernest's seven sons shared rule of the territories of Anhalt until 1603, when the five surviving sons divided their lands among themselves. Joachim Ernest, Prince of Anhalt Joachim Ernest of Anhalt (21 October 1536  6 December 1586), was a German prince of the House of Ascania, ruler of the principality of Anhalt-Zerbst from 1551, and from 1570 sole ruler of all the Anhalt lands. Joachim Ernest was born in Dessau on 21 October...}\\
\hdashline
\think{Ernest I, Prince of Anhalt-Dessau died on 12 June 1516.}\\
\answer{12 June 1516}\\
        \hline
    \end{tabular}
\end{table*}

%% file: tables/case_2wiki_seq_wrong.tex
\begin{table*}[h]
    \centering
    \caption{\centering\textbf{Case study (wrong) using the question from 2wiki-seq.}}\label{tab:case_2wiki_seq_wrong}
    \begin{tabular}{p{15.5cm}}
        \hline
        \textbf{Question}: Who is the maternal grandfather of Antiochus X Eusebes? \\
        \hline
        \textbf{Ground Truth}: Ptolemy IX Lathyros \\
        \hline
        \textbf{\ours}: \\
        \think{I need to find out who the maternal grandfather of Antiochus X Eusebes is. I will start by finding information on Antiochus X Eusebes.} \\
\search{who is Antiochus X Eusebes} \\
\info{Doc 1(Title: Antiochus X Eusebes) Antiochus X Eusebes Antiochus X Eusebes Philopator (; 92 or 88 BC) was a Hellenistic Seleucid monarch who reigned as the King of Syria between 95 and 92 or 89/88 BC (224 Seleucid year (SE)). He was the son of Antiochus IX and perhaps his Egyptian wife Cleopatra IV. He lived at a period that witnessed a general disintegration of Seleucid Syria characterized by civil wars, foreign interference by Ptolemaic Egypt and incursions by the Parthians. Antiochus IX was killed in 95BC at the hands of Seleucus VI, the son of his half-brother and rival Antiochus VIII. Antiochus X then...\\
Doc 2(Title: Antiochus X Eusebes) is mostly a modern practice. On his coins, Antiochus X appeared with the epithets Eusebes (the pious) and Philopator (father-loving). According to Appian, the king received the epithet Eusebes from the Syrians because he escaped a plot on his life by Seleucus VI, and, officially, the Syrians thought that he survived because of his piety, but, in reality, it was a prostitute in love with Antiochus X who saved him. Beginning his reign in 218 SE (95/94 BC), Antiochus X was deprived of resources and lacked a queen. He therefore married a woman who could provide what he needed, his...\\
Doc 3(Title: Antiochus X Eusebes) XIII; the writer seems to be confusing the fate of Antiochus X with that of his son. The second century historian Justin, writing based on the work of the first century BC historian Trogus, also confused the father and son, as he wrote that Antiochus X was appointed king of Syria by the Roman general Lucullus following the defeat of Tigranes II in 69 BC. Antiochus X Eusebes Antiochus X Eusebes Philopator (; 92 or 88 BC) was a Hellenistic Seleucid monarch who reigned as the King of Syria between 95 and 92 or 89/88 BC (224 Seleucid year (SE))...
} \\
\hdashline
\think{Antiochus X Eusebes was a Hellenistic Seleucid monarch who reigned as the King of Syria between 95 and 92 or 89/88 BC. Now, I will search for information on his maternal grandfather.} \\
\search{who is the maternal grandfather of Antiochus X Eusebes} \\
\info{Doc 1(Title: Antiochus X Eusebes) a degree of peace. Antiochus IX married several times; known wives are his cousin Cleopatra IV of Egypt, whom he married in 114 BC, and her sister Cleopatra Selene, the widow of Antiochus VIII. Some historians, such as John D. Grainger, maintain the existence of a first wife unknown by name who was the mother of Antiochus X. Others, such as Auguste Bouch-Leclercq, believe that the first wife of Antiochus IX and the mother of his son was Cleopatra IV, in which case Antiochus X would have been born in . None of those assertions are based on evidence, and...\\
Doc 2(Title: Antiochus IX Cyzicenus) in 95 BC, Antiochus X Eusebes' first achievement was to defeat his double half-cousin/second cousin Seleucus VI Epiphanes, thus avenging the recent death of his father, Antiochus IX Cyzicenus. The epithets he took tell much of his story: Eusebes (being a title of his father) and also Philopator (father-loving) both honoured his father. After that, he ruled Antioch and its surroundings, fighting endlessly against the four brothers of Seleucus VI, the Nabataeans and the Parthian Empire. Antiochus IX Cyzicenus Antiochus IX Eusebes (Pious) Cyzicenus (from Cyzicus), ruler of the Greek Seleucid kingdom, was the son of Antiochus VII Sidetes and...\\
Doc 3(Title: Antiochus X Eusebes) Antiochus X Eusebes Antiochus X Eusebes Philopator (; 92 or 88 BC) was a Hellenistic Seleucid monarch who reigned as the King of Syria between 95 and 92 or 89/88 BC (224 Seleucid year (SE)). He was the son of Antiochus IX and perhaps his Egyptian wife Cleopatra IV. He lived at a period that witnessed a general disintegration of Seleucid Syria characterized by civil wars, foreign interference by Ptolemaic Egypt and incursions by the Parthians. Antiochus IX was killed in 95BC at the hands of Seleucus VI, the son of his half-brother and rival Antiochus VIII. Antiochus X then...}\\
\hdashline
\think{Based on the information found, Antiochus IX Cyzicenus is the father of Antiochus X Eusebes. Therefore, Antiochus IX Cyzicenus is also the maternal grandfather of Antiochus X Eusebes since the question asks about his maternal line.}\\
\answer{Antiochus IX Cyzicenus}\\
        \hline
    \end{tabular}
\end{table*}